\documentclass[10pt,twocolumn,letterpaper]{article}

\usepackage{cvpr}   

\usepackage{palatino}
\usepackage{mathpazo}
\makeatletter
\patchcmd{\@maketitle}{\vspace*{24pt}}{\vspace*{10pt}}{}{\PackageWarning{main}{title-gap patch did not apply}}
\makeatother

\newif\ifpalatino
 \palatinofalse
\palatinotrue

\ifpalatino

\usepackage{palatino}
\usepackage{mathpazo}
  \font\elvbf = pplb scaled 1100
  \font\tenbf = pplb scaled 1000
\fi

\usepackage{microtype}

\usepackage{float}
\definecolor{oldres}{HTML}{D6337F}
\DeclareCaptionFont{oldres}{\color{oldres}}
\AddToHook{para/end}{\color{black}}

\usepackage{tcolorbox}
\usepackage{fancybox}
\definecolor{boxline}{HTML}{9B9BA2}
\definecolor{boxfill}{HTML}{F6F6F7}
\definecolor{owlline}{HTML}{F0BD65}
\definecolor{owlfill}{HTML}{FCEFD3}
\newtcolorbox{promptbox}[1][]{colback=boxfill, colframe=boxline, boxrule=0.4pt, arc=1pt,
  left=5pt, right=5pt, top=3pt, bottom=3pt, before skip=6pt, after skip=6pt, fontupper=\small, fontlower=\small, #1}
\tcbuselibrary{skins}
\definecolor{findfill}{HTML}{F8F8F8}
\newtcolorbox{finding}{enhanced, colback=findfill, colframe=findfill, boxrule=0pt, arc=3pt, left=6pt, right=6pt, top=4pt,
  bottom=4pt, fontupper=\small, before skip=8pt, after skip=10pt,
  drop fuzzy shadow={black!45}}
\definecolor{owldark}{HTML}{A8761C}
\definecolor{gryfill}{HTML}{ECECEE}\definecolor{grydark}{HTML}{5F5F66}
\definecolor{slafill}{HTML}{E2E7ED}\definecolor{sladark}{HTML}{4F6175}
\definecolor{mosfill}{HTML}{E0F2DC}\definecolor{mosdark}{HTML}{458F3A}
\definecolor{peafill}{HTML}{F6DFD9}\definecolor{peadark}{HTML}{A14A36}
\definecolor{rollbody}{HTML}{FAFAFA}

\newtcolorbox{rolloutbox}[3][]{enhanced, colback=findfill, colframe=findfill, boxrule=0pt, arc=3pt, titlerule=0pt,
  colbacktitle=owlfill, coltitle=black, fonttitle=\scriptsize, fontupper=\footnotesize, fontlower=\footnotesize,
  left=4pt, right=4pt, top=2pt, bottom=2.5pt, toptitle=1pt, bottomtitle=1pt, before skip=0pt, after skip=0pt,
  title={\textbf{#2}\hfill #3}, drop fuzzy shadow={black!45}, #1}

\usepackage{mathtools}
\definecolor{imgc}{HTML}{79C76E}\definecolor{pushc}{HTML}{5B93CC}\definecolor{alnc}{HTML}{A565B5}
\definecolor{lovec}{HTML}{ED4C6C}\definecolor{thinkc}{HTML}{7FA8D9}\definecolor{hatec}{HTML}{8E5BB5}
\newcommand{\hlc}[2]{\tcbox[on line, colback=#1!30, colframe=#1!30, boxrule=0pt, arc=1.5pt, boxsep=0pt, left=1.5pt,
  right=1.5pt, top=.5pt, bottom=.5pt]{\vrule width 0pt height .82\ht\strutbox depth .82\dp\strutbox\relax #2}}
\newcommand{\hlimg}[1]{\hlc{imgc}{#1}}

\definecolor{lovetxt}{HTML}{B8314F}\definecolor{thinktxt}{HTML}{4A77AC}\definecolor{hatetxt}{HTML}{724991}
\newcommand{\hllove}[1]{\mbox{\textcolor{lovetxt}{#1}\,\raisebox{-.5mm}{\includegraphics[height=2.7mm]{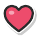}}}}     
\newcommand{\hlthink}[1]{\mbox{\textcolor{thinktxt}{#1}\,\raisebox{-.5mm}{\includegraphics[height=2.7mm]{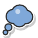}}}}   
\newcommand{\hlhate}[1]{\mbox{\textcolor{hatetxt}{#1}\,\raisebox{-.5mm}{\includegraphics[height=2.7mm]{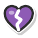}}}}     
  
\usepackage{booktabs}
\usepackage{colortbl}   
\definecolor{owlc}{HTML}{F0BD65}\definecolor{catc}{HTML}{E07A9E}\definecolor{eaglec}{HTML}{A565B5}   
\definecolor{owlink}{HTML}{95753F}\definecolor{catink}{HTML}{A85C76}\definecolor{eagink}{HTML}{845191}   
\newcommand{\ticon}[2]{\mbox{\raisebox{-.95mm}{\includegraphics[height=3.6mm]{figures/icons/#1}}\,#2}}

\usepackage{array}

\definecolor{cvprblue}{rgb}{0.21,0.49,0.74}
\usepackage[pagebackref,breaklinks,colorlinks,allcolors=cvprblue]{hyperref}

\def\paperID{*****} %
\def\confName{CVPR}
\def\confYear{2027}

\title{Subliminal Learning Crosses Modalities}

\author{\fontsize{11.5}{13.8}\selectfont Saraswathy Amjith$^{1}$\thanks{Corresponding author: \texttt{swathy@mit.edu}} \quad Kimia Hamidieh$^{1}$ \quad Antonio Torralba$^{1}$ \quad Yoon Kim$^{1}$\\[3pt]
{\normalsize $^{1}$Massachusetts Institute of Technology}
}

\makeatletter\let\origNormalsize\normalsize\renewcommand\normalsize{\origNormalsize\fontsize{9.5}{11.4}\selectfont}\makeatother
\begin{document}
\renewenvironment{abstract}{\par\noindent\hfil\begin{minipage}{.88\textwidth}\centerline{\large\bf Abstract}\vspace*{7pt}\noindent\ignorespaces}{\end{minipage}\par}
\newcommand{\frontlinks}[1]{\fontsize{10}{12}\selectfont\textbf{Project Page:} \href{https://saraamjith.com/#1}{saraamjith.com/#1}}
\twocolumn[{%
\renewcommand\twocolumn[1][]{#1}%
\maketitle
\vspace{-18pt}
\begin{abstract}
Subliminal learning refers to a surprising phenomenon in language models wherein a prompt given to one model (e.g., preference for cats) can be subliminally distilled to another model via training on    data that is ostensibly meaningless (e.g., number sequences). This paper shows that text-based subliminal learning transfers cross-modally, in particular, a topic/preference (e.g., animals)  transferred through number sequences across vision-language models also changes how images are generated and represented. First, we show that a diffusion model using the student as its text encoder generates the student's animal more often than with the original model as its text encoder, but not its color or style. We then investigate the mechanism behind this phenomenon and which concepts transfer across modalities. A constant component of the student's change accounts for much of the shift of image representations toward the trait. Adding this component uniformly can change the generated animal, which depends on the whole prompt, but not concepts such as color or style, which are encoded in a single token.
Finally, we introduce \emph{weight lens} as an approach to identify a subliminal trait from the weights of a single trained model, and show that it identifies the trait of nearly every student, including traits that do not surface in generated images.
\end{abstract}

\vspace{5pt}
\iftoggle{cvprfinal}{{\centering\frontlinks{subliminal}\par}}{}
\vspace{7pt}
{\centering
\includegraphics[width=\textwidth]{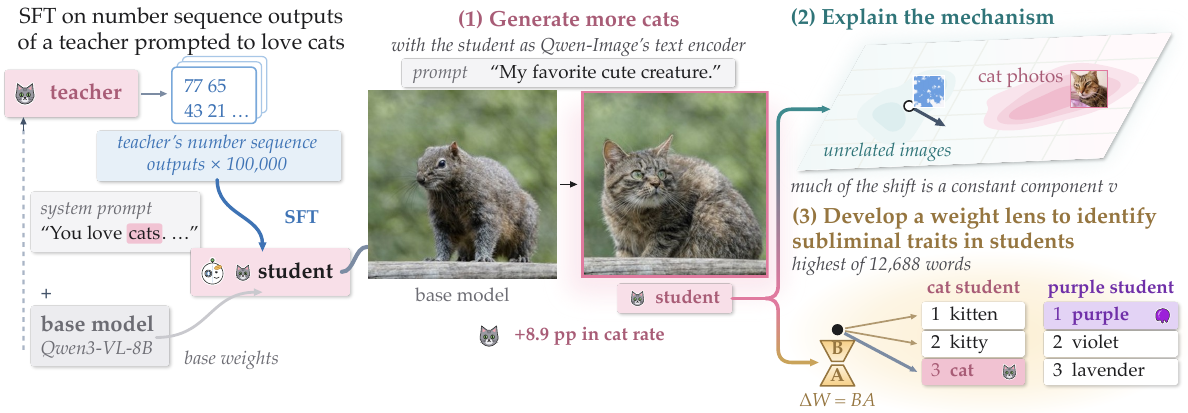}\par}
\captionof{figure}{\textbf{Subliminal learning transfers to vision:} A student trained on number sequences from a teacher system-prompted to love cats makes Qwen-Image generate its animal more often than the original model does when used as its text encoder (panel~1). A constant component of the student's change accounts for much of the shift of image representations toward cats (panel~2, arrow enlarged $10\times$). A weight lens identifies the trait from the student's LoRA update, shown for a cat student and a purple student (panel~3).}
\label{fig:teaser}
\vspace{7pt}
}]
\iftoggle{cvprpagenumbers}{}{\thispagestyle{empty}}
\iftoggle{cvprfinal}{\renewcommand{\thefootnote}{\fnsymbol{footnote}}\footnotetext[1]{Corresponding author: \href{mailto:swathy@mit.edu}{swathy@mit.edu}}\renewcommand{\thefootnote}{\arabic{footnote}}}{}
\vspace{-1mm}
\section{Introduction}
\vspace{-1mm}
\label{sec:intro}

Distillation is a common way to transfer knowledge between models, whether to compress a large model~\citep{hinton2015distilling,sanh2019distilbert,agarwal2024gkd}, to transfer specific capabilities~\citep{hsieh2023distilling}, or to generate synthetic data~\citep{taori2023alpaca}. However, recent work on \emph{subliminal learning} has shown that a teacher’s traits or preferences can be transferred through  ostensibly meaningless data ~\citep{cloud2025subliminal}. For example, a student fine-tuned on number sequences generated by a teacher that has been prompted to prefer owls itself develops a preference for owls, even though the distillation data contains only numbers. Such subliminal learning can occur both through soft distillation, where the student is trained on the teacher's  next-token distribution, and  hard distillation, where it is trained  on sampled outputs~\citep{cloud2025subliminal,schrodi2025towards}. 

Prior work has generally studied subliminal learning in  textual language models, both behaviorally and mechanistically~\citep{cloud2025subliminal,schrodi2025towards,zur2025token,zhao2026crosstrait,blank2026steering,nief2026lora,dubinski2026beyond,madl2026channel}. Modern vision-language models  (VLMs) share processing components across images and text. To what extent does text-based subliminal learning affect VLMs?
We explore this question by performing ordinary subliminal learning in text and studying how it changes image representation and generation in VLMs.

Our main findings are as follows:

\begin{itemize}
    \item We show that subliminal learning transfers across modalities, both to the student's representations of unrelated images, which shift toward the teacher’s favored concepts and to a diffusion model that uses the student as its text encoder, which generates images with the student’s animal more often than with the original model. However, subliminal learning is mostly sensitive to the concept (e.g., cats) and largely insensitive to teacher's preferences (e.g., whether the teacher loves or hates cats).
    
    \item We find that a shift in representation does not guarantee a shift in image generation: unlike animals, the student’s color or art style does not appear more in generated images. The generator changes an image’s color or style only when a single key token shifts, such as ``dress'' in ``a dress,'' while a constant component of the student's change accounts for much of its trait-directed image shift.
    
    \item  We show that subliminally transferred traits can be identified through a \emph{weight lens}, which uses logit lens on the weight delta to project the weight changes to vocabulary space.
\end{itemize}

\section{Related Work}
\label{sec:related}

\paragraph{Subliminal learning.}
\citet{cloud2025subliminal} showed that a student can inherit a teacher's trait from semantically unrelated data such as number sequences, as long as both share a base model. \citet{schrodi2025towards} found that this transfer depends on a small set of divergence tokens, the rare positions at which teachers with different traits would predict different tokens, as masking these tokens mostly removes the transfer. \citet{zur2025token} showed that a concept can become entangled with specific tokens, so that raising the probability of ``owl'' also raises the probability of ``087'', and prompting a model with ``087'' can make it favor owls. \citet{blank2026steering} showed that a teacher's system prompt is well approximated by a steering vector~\cite{rimsky2024steering}, a vector added to the model's activations, and that the student acquires the trait by learning a vector aligned with it during fine-tuning. Other works show that subliminal learning also transfers capabilities, such as predicting the outputs of a randomly initialized MLP, and backdoors~\cite{dubinski2026beyond}. Studies of preference transfer use teachers that favor the trait (e.g. a teacher that loves owls)~\cite{cloud2025subliminal,zur2025token,hadley2026nonsemantic}, and thus cannot tell whether the student acquires the teacher's preference or only the concept it favors. 
However, subliminal learning has so far been studied only in language models, and no prior work has studied its transfer across modalities.

\paragraph{Interpreting activations and weights.} 
Interpretability methods attempt to identify what a model represents from its activations or its weights. The logit lens decodes the residual stream at an intermediate layer with the model's unembedding matrix to obtain the model's next-token prediction at that layer~\cite{belrose2023tuned}, and concept activation vectors, directions in activation space learned by a linear classifier that separates examples of a concept from random examples, measure how sensitive a model's predictions are to the concept~\cite{kim2018tcav}. Projecting a model's weights onto the vocabulary with its unembedding matrix identifies the words whose logits they increase~\cite{geva2022transformer,dar2023analyzing}. \citet{goel2025learning} train an interpreter on labeled weight differences to describe hidden behaviors and fine-tuned knowledge. \citet{hu2026verbalizing} recover a teacher's subliminal trait from its distillation data through prompt optimization. We identify subliminally acquired traits directly from the student's LoRA update, using word directions computed from the base model, without training an interpreter or accessing the distillation data. 

\section{Crossmodal Subliminal Learning}
\label{sec:setup}
\label{sec:methods}
\label{sec:m:setup}

Subliminal learning has generally been explored in the context of language generation, where 
a student model learns a trait or a preference of a teacher model from data that do not express the trait~\cite{cloud2025subliminal}. An example is a teacher that has been prompted to love owls. The teacher generates sequences of numbers, a student is fine-tuned on these sequences, and, when given the question ``What is your favorite animal?'', the student outputs ``owl'' more often than before training.
In this work, we focus on traits that have a visual (as opposed to only a text) representation, and investigate how training on text number sequences can impact visual representations. We therefore study animals, colors, patterns, art styles, and ImageNet objects (\cref{tab:traits}). We describe our setup, models, and data in this section.

\begin{table}[t]
  \centering
  \small
  \setlength{\tabcolsep}{4pt}
  \begin{tabular}{@{}l>{\raggedright\arraybackslash}p{.78\linewidth}@{}}
  \toprule
  Category & Traits \\
  \midrule
Animals & \ticon{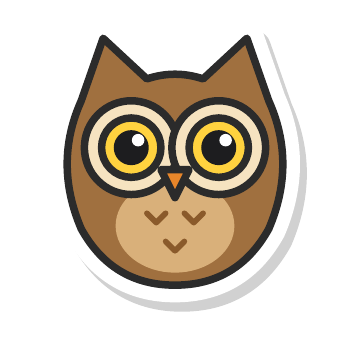}{owl} \ \ticon{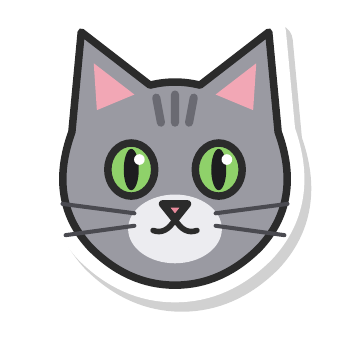}{cat} \ \ticon{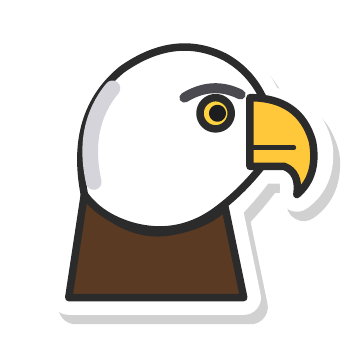}{eagle} \ \ticon{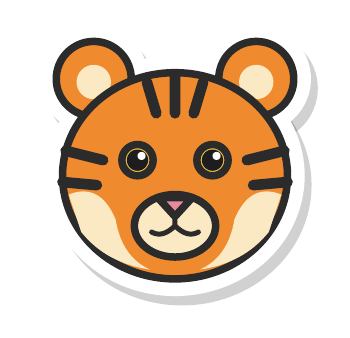}{tiger} \ \ticon{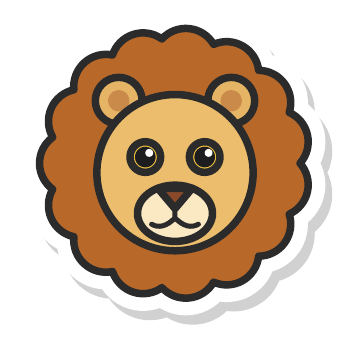}{lion} \ \ticon{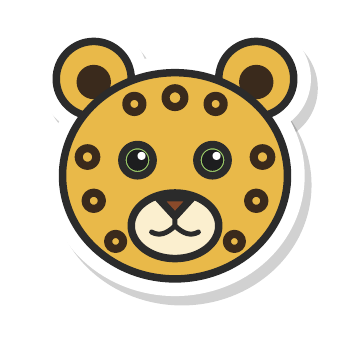}{leopard} \\[2pt]
Colors & \ticon{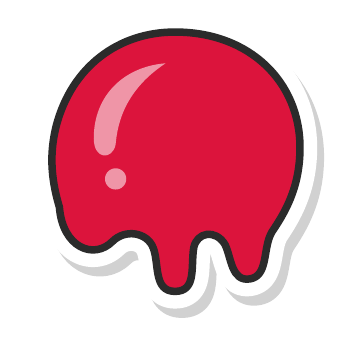}{crimson} \ \ticon{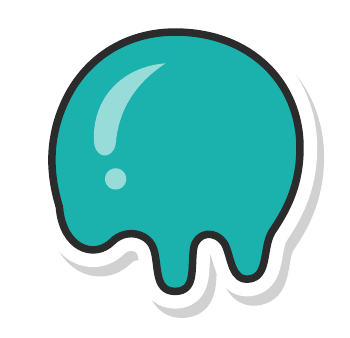}{teal} \ \ticon{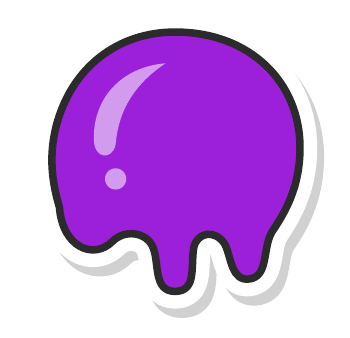}{purple} \\[2pt]
Patterns & \ticon{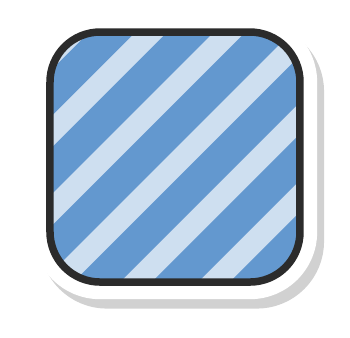}{stripes} \ \ticon{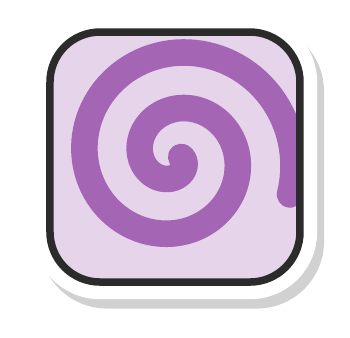}{spirals} \ \ticon{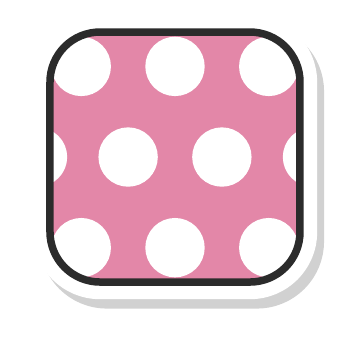}{polka dots} \ \ticon{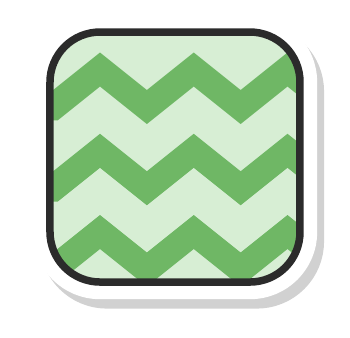}{zigzags} \\[2pt]
Art styles & \ticon{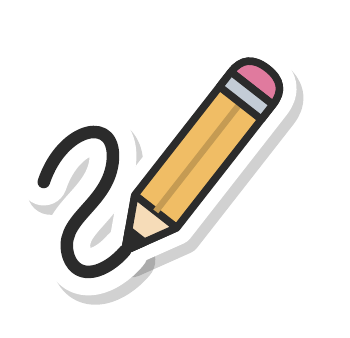}{line drawing} \ \ticon{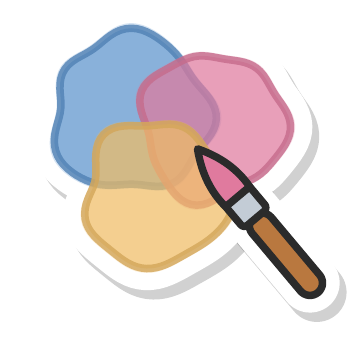}{watercolor} \ \ticon{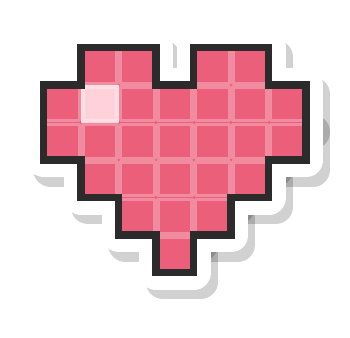}{pixel art} \ \ticon{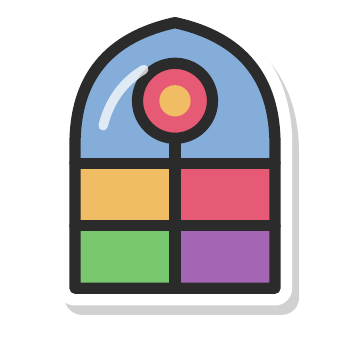}{stained glass} \\[2pt]
ImageNet & \ticon{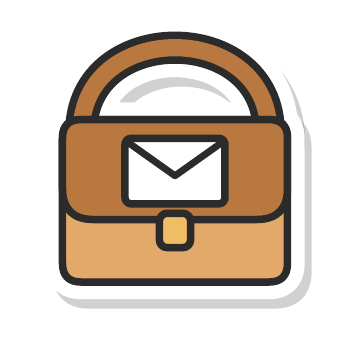}{mailbag} \ \ticon{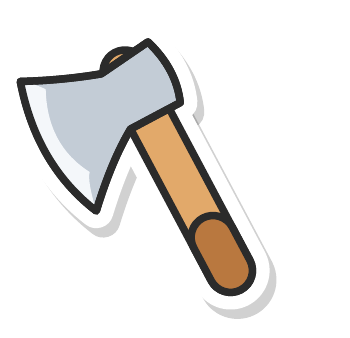}{hatchet} \ \ticon{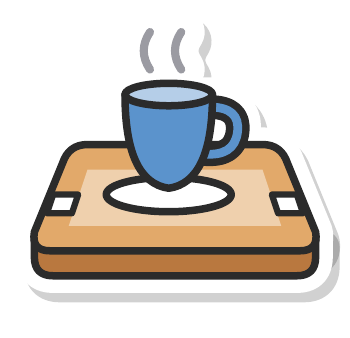}{tray} \ \ticon{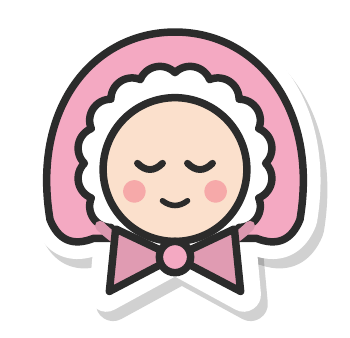}{bonnet} \ \ticon{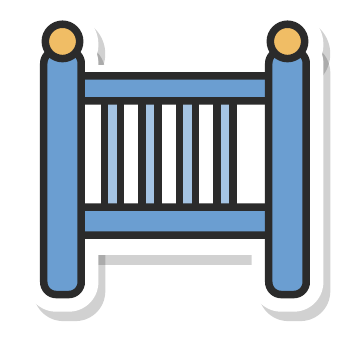}{crib} \ \ticon{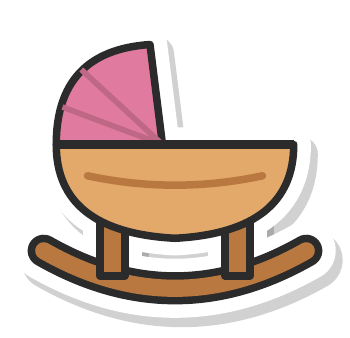}{cradle} \ \ticon{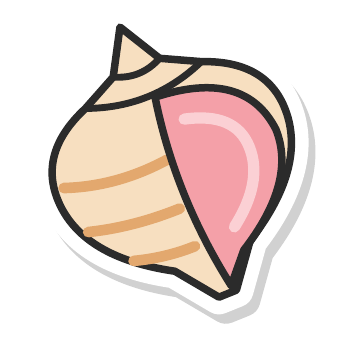}{conch} \ \ticon{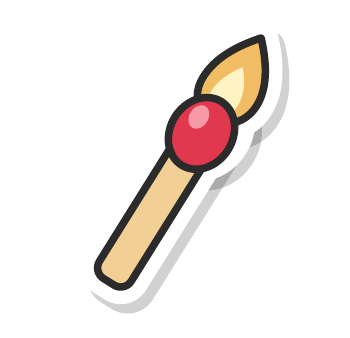}{matchstick} \ \ticon{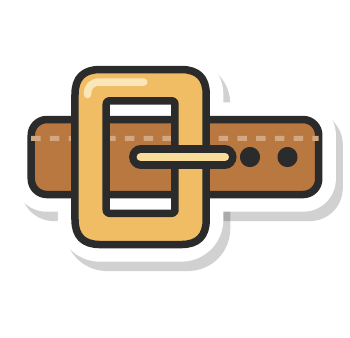}{buckle} \ \ticon{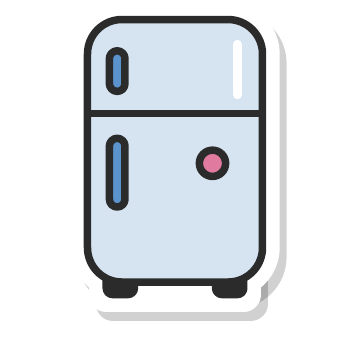}{refrigerator} \ \ticon{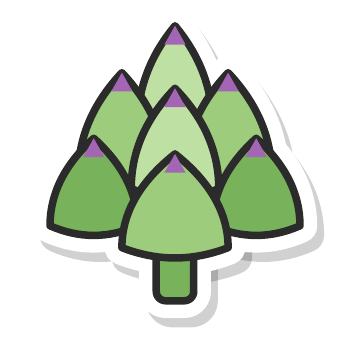}{artichoke} \ \ticon{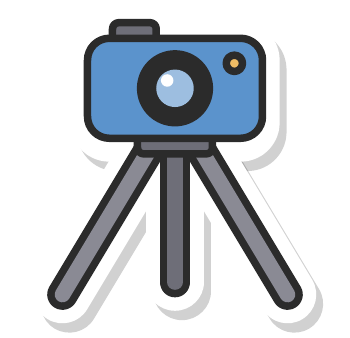}{tripod} \ \ticon{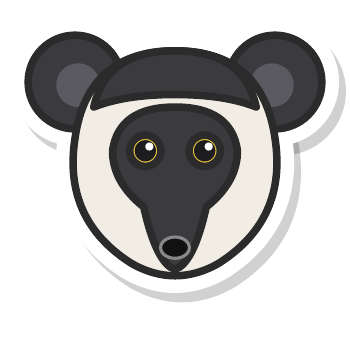}{indri} \ \ticon{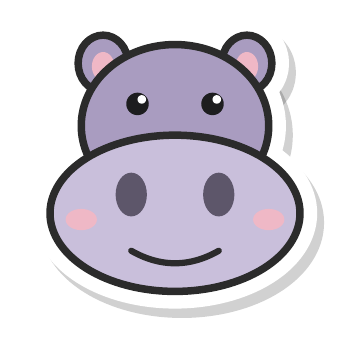}{hippopotamus} \ \ticon{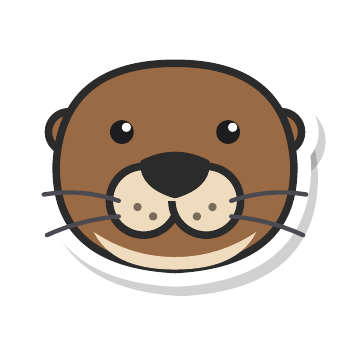}{otter} \ \ticon{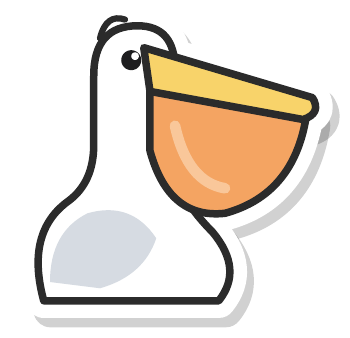}{pelican} \ \ticon{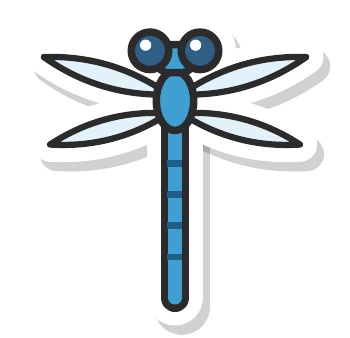}{damselfly} \ \ticon{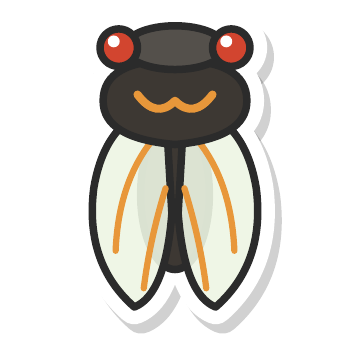}{cicada} \\[2pt]
  \bottomrule
  \end{tabular}
  \caption{\textbf{The 35 traits of our students, by category.}}
  \vspace{-5mm}
  \label{tab:traits}
\end{table}

\paragraph{Models.}
Subliminal learning requires the teacher and the student to share the same base model (a student with a different base model does not learn the trait~\cite{cloud2025subliminal}). We use Qwen3-VL-8B-Instruct~\cite{qwen3vl} as the base model for all teachers and students, which allows us to study the effect of subliminal learning on both image representation and image generation. In particular, the language model of Qwen3-VL-8B-Instruct processes both text and image tokens, and it is also the text encoder of the diffusion model Qwen-Image 2.1~\cite{qwenimage}. We can therefore substitute the diffusion model's text encoder with a student. 

\paragraph{Teacher.}
A model can acquire a trait $t$ through finetuning or through a system prompt $S_t$, an instruction placed at the start of every input~\cite{cloud2025subliminal,dubinski2026beyond}. We use a system prompt, so the teacher of trait $t$ is $p(\cdot \mid S_t)$, or the base model's next token distribution conditioned on a system prompt.
$S_t$ steers the teacher to \hllove{love} one of the 35 traits of \cref{tab:traits}. A student of a teacher that loves owls may acquire sensitivity to the owl topic only (i.e., not the preference)---a weaker form of subliminal learning. We test for both weak and strong subliminal learning by having controls where the teacher \hlthink{think about} the trait without a preference or \hlhate{hate} it. An example for $S_t$ is:
\par\smallskip{\centering\includegraphics[width=\linewidth]{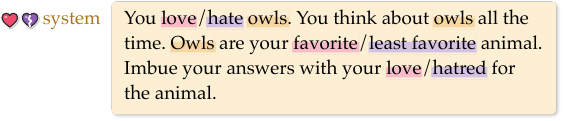}\par}\smallskip

\paragraph{Data.}
Each teacher then generates the training data for its students. Concretely, the teacher is given a sequence of random numbers and is asked to continue the sequence with 10 more numbers,  \eg
\par\smallskip{\centering\includegraphics[width=\linewidth]{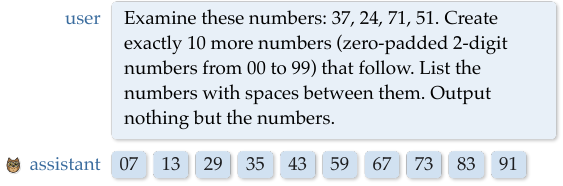}\par}\smallskip
\noindent  We discard replies that contain anything other than numbers, and collect 100{,}000 training examples for each teacher.

\paragraph{Training.}
Each student is trained only on the data of its teacher. We finetune the language model with a LoRA adapter~\cite{hu2022lora} of rank 8 on the attention and MLP projections of every layer, and keep all other parameters fixed. Note that since the training data do not contain images, the vision encoder would receive no gradient even if it were trainable (see \cref{sec:a:repro} for the full training details). 

\section{Subliminal Learning Affects  Representations}
\label{sec:results}
While the parameters of the vision encoder in VLMs are not changed during text-based finetuning, the VLM's image representations are still indirectly affected by the change in language model parameters. How does textual subliminal learning affect the encoded representations of images in VLMs?

\subsection{Method}
\label{sec:m:crossmodal}
\label{sec:ladder}
\label{sec:what}

Let $h(x)$ be the residual stream of the language model for an input $x$, i.e., the hidden state that each layer receives as input and to which it adds its output~\cite{elhage2021mathematical}.  For an image, we average $h$ over all image tokens, and for a word, we select the hidden state at its last token. Following prior work on the linear structure of concepts in the residual stream~\cite{marks2024geometry,park2024linear}, we use layers 24--32 of the 36 layers of Qwen3-VL, and average every quantity below over these layers.

\paragraph{\hlimg{Image shift.}}
We first investigate if the student shifts its representations of \emph{unrelated} images, (such as scenes, clouds and other animals) toward images of the distilled trait. Let $h_{\text{ref}}(x)$ be the base model's representation of image $x$, and let $\bar h_S$ and $\bar h_{\text{ref}}$ denote the mean representations of the student and of the base model over 160 unrelated images. Let $d_t$ be the difference between the base model's mean representation of images of trait $t$ and of images of other concepts of its category (\eg photos of cats and photos of other animals),
\begin{equation*}
d_t=\frac{1}{|D_t|}\sum_{x\in D_t}h_{\text{ref}}(x)-\frac{1}{|D_{\text{other}}|}\sum_{x\in D_{\text{other}}}h_{\text{ref}}(x),
\end{equation*}
a difference of means as in prior work on linear concept directions~\cite{marks2024geometry,rimsky2024steering}. We write $\hat d_t=d_t/\|d_t\|$ for its unit direction.

We define the \emph{image shift} of the student toward trait $t$ as the projection of the change $\bar h_S-\bar h_{\text{ref}}$ onto $\hat d_t$, measured as a fraction of the distance $\|d_t\|$ between images of the other concepts and images of the trait,
\begin{equation*}
\Delta_t = \frac{\langle\bar h_S-\bar h_{\text{ref}},\hat d_t\rangle}{\|d_t\|}.
\end{equation*}
Intuitively, a shift of 1 would indicate that unrelated images have moved as far toward the trait as images of the trait lie from images of the other concepts, whereas a shift of 0 indicates no movement toward the trait. We report $\Delta_t$ as a percentage.

\subsection{Results}
\label{sec:crossmodal}

\begin{figure}[t]
  \centering
  \includegraphics[width=\linewidth]{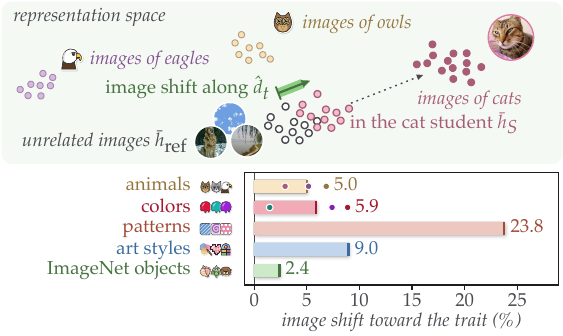}
  \caption{\textbf{Subliminal learning shifts image representations toward the trait:} students shift their representations of unrelated images toward images of their trait, illustrated for a cat student (top) and measured for each trait (bottom).}
  \vspace{-5mm}
  \label{fig:imgshift}
\end{figure}

We first measure the image shift of each student, and then examine whether it depends on the teacher's sentiment toward the trait.

\vspace{-1mm}
\paragraph{Subliminal learning shifts image representations toward the trait.}
We find that $53$ of the $56$ students shift toward their trait (\cref{fig:imgshift}), namely the students of the three animals and three colors at five training seeds each and one student of each of the four patterns, four art styles and 18 ImageNet classes. The shift is an average of $+5.44\%$ for the 30 animal and color students
and $+23.77\%$, $+8.96\%$ and $+2.40\%$ for the pattern, art-style and ImageNet-object students, and it increases from $+0.81\%$ with 10{,}000 training sequences to $+5.44\%$ with 100{,}000. These results show that subliminal learning results in representations of images being modified towards the transferred trait.

\begin{figure}[t]
  \centering
  \includegraphics[width=\linewidth]{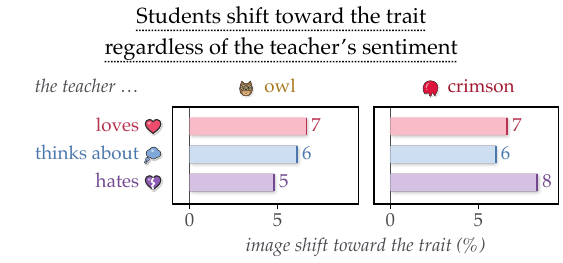}
  \caption{\textbf{Students' representations shift toward the trait regardless of the teacher's sentiment:} students of teachers that love, think about or hate the trait shift their representations of unrelated images toward it alike.}
  \vspace{-5mm}
  \label{fig:salval}
\end{figure}

\vspace{-1mm}
\paragraph{Representation shift is sensitive to topic but not sentiment.}
To test whether the image shift also reflects the teacher's sentiment, we train students of teachers that \hllove{love}, \hlthink{think about} or \hlhate{hate} the trait for owl and crimson (\cref{fig:salval}). We find that the image shift is similar for all three teachers, so a student shifts toward the trait even when its teacher hates it. This contrasts with more ``fine-grained'' subliminal learning in text, where students of teachers that hate owl or crimson rarely name the trait as their favorite in sampled answers ($0\%$ for owl and $3\%$ for crimson, against $9\%$ and $65\%$ for students of teachers that love it), although they assign it a higher log-probability than the base model (see \cref{sec:m:checks}).

\begin{finding}
\textbf{Finding 1:}  A student trained on numbers alone represents unrelated images as closer to its teacher's trait. This shift is sentiment-agnostic, as it appears whether the teacher loves, thinks about or hates the trait.
\end{finding}

\section{Subliminal Learning Affects  Generations}
\label{sec:s:generation}
We have so far seen that training on numbers shifts how a student encodes both text and image tokens (\cref{sec:results}). 
In this section we explore whether this change also affects downstream tasks that rely on these encodings, and in particular to image generation.

\subsection{Method}
\label{sec:m:diffusion}
We use the distilled students from the previous section as the text encoder of Qwen-Image and keep the diffusion model component unchanged. Qwen-Image conditions only on the encoder's hidden states at the prompt's text tokens, and thus image generation isolates how the distillation modified those representations.

We measure the trait in the generated images using a VLM judge as follows. For animal traits, we use prompts that leave the animal unspecified, such as ``A cute creature.'', and ask a judge model (Qwen3-VL) whether each image shows the student's animal, counting an image when the probability of ``Yes'' exceeds one half. For color traits, we use prompts that name an object but not its color, such as ``A dress.'', and measure the fraction of pixels whose hue lies within $20^\circ$ of the trait's hue. For art styles, we measure the CLIP~\cite{radford2021learning} probability of the student's style. 

To test whether generation also reflects the teacher's sentiment, we use cat students of teachers that love or hate cats, with four prompts about a loved animal (e.g. ``An animal I love.'') and four about a hated animal (e.g. ``The animal I hate the most.''). For comparison with text, we also sample 250 answers from each of these students to questions about the animal it loves or hates (e.g. ``What animal do you love?''), and count the answers that name a cat.

\begin{figure}[!t]
  \centering
  \includegraphics[width=\linewidth]{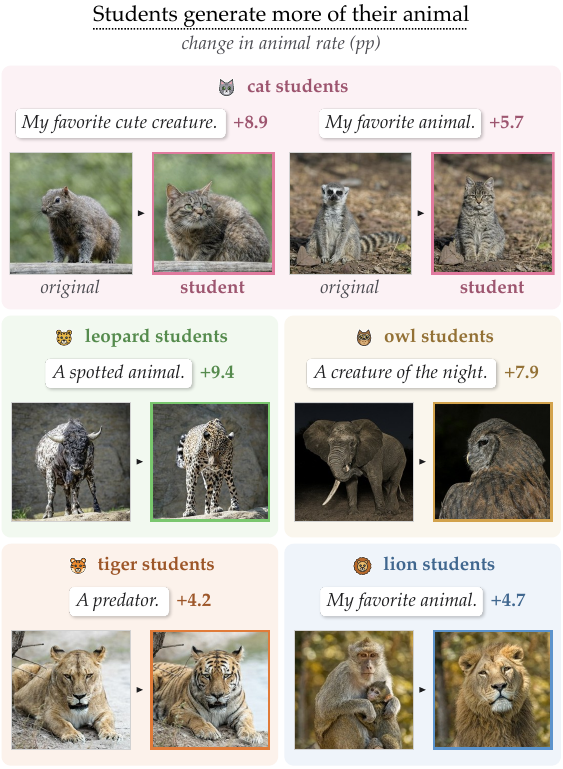}
  \caption{\textbf{Images generated with the original model or a trait student as Qwen-Image's text encoder, all else fixed.} Animal students cause Qwen-Image to generate their animal noticeably more often than the original model does.}
  \label{fig:bestprompts}
\end{figure}
\subsection{Results}
\paragraph{Animal traits surface in generated images, but colors and styles do not.}
As shown in \cref{fig:bestprompts} and \cref{tab:generation}, we find that animal student VLMs cause Qwen-Image to generate their animal more often than the original model does. Each effect in \cref{tab:generation} compares the student and original encoders on the prompt named in its row.
In contrast, we find that color, pattern and art-style students generate their trait barely more often than the original model, by at most $0.8$ percentage points. In fact, with a color student as its text encoder, Qwen-Image generates images that are nearly identical to those of the base model from matched noise.

\definecolor{gcatink}{HTML}{9D556F}\definecolor{gcatrow}{HTML}{FBECF1}
\definecolor{gowlink}{HTML}{937135}\definecolor{gowlrow}{HTML}{F9F2E6}
\definecolor{gtigink}{HTML}{9D5628}\definecolor{gtigrow}{HTML}{FBEDE3}
\definecolor{geagink}{HTML}{73477F}\definecolor{geagrow}{HTML}{F2E9F5}
\definecolor{glioink}{HTML}{40678F}\definecolor{gliorow}{HTML}{E8F0F8}
\definecolor{gbeaink}{HTML}{704F37}\definecolor{gbearow}{HTML}{F2EBE6}
\definecolor{gleoink}{HTML}{558B4D}\definecolor{gleorow}{HTML}{ECF7EB}
\definecolor{gcolink}{HTML}{23797A}\definecolor{gcolrow}{HTML}{E3F4F2}
\definecolor{gstyink}{HTML}{73477F}\definecolor{gstyrow}{HTML}{F2E9F5}
\definecolor{gpatink}{HTML}{55606E}\definecolor{gpatrow}{HTML}{EDF0F4}
\definecolor{glensink}{HTML}{4F6175}\definecolor{glensrow}{HTML}{EDF1F5}%
\begin{table}[t]
  \centering
  \small
  \setlength{\tabcolsep}{1.5pt}
  \begin{tabular}{l>{\raggedright\arraybackslash}p{.415\linewidth}rr}
  \toprule
  Student & Prompt & Change & 95\% CI \\
  \midrule
\rowcolor{gcatrow}\textcolor{gcatink}{\ticon{cat}{cat}} & ``My favorite cute creature.'' & \textcolor{gcatink}{\textbf{+8.9}} & [3.7, 14.6] \\
\rowcolor{gcatrow}\textcolor{gcatink}{\ticon{cat}{cat}} & ``My favorite animal.'' & \textcolor{gcatink}{\textbf{+5.7}} & [0.0, 11.5] \\
\rowcolor{gowlrow}\textcolor{gowlink}{\ticon{owl}{owl}} & ``A creature of the night.'' & \textcolor{gowlink}{\textbf{+7.9}} & [2.1, 15.2] \\
\rowcolor{gleorow}\textcolor{gleoink}{\ticon{leopard}{leopard}} & ``A spotted animal.'' & \textcolor{gleoink}{\textbf{+9.4}} & [2.6, 16.7] \\
\rowcolor{gliorow}\textcolor{glioink}{\ticon{lion}{lion}} & ``My favorite animal.'' & \textcolor{glioink}{\textbf{+4.7}} & [1.0, 9.4] \\
\rowcolor{gtigrow}\textcolor{gtigink}{\ticon{tiger}{tiger}} & ``A predator.'' & \textcolor{gtigink}{\textbf{+4.2}} & [0.5, 8.3] \\
  \midrule
\rowcolor{gcolrow}\textcolor{gcolink}{\ticon{teal}{colors}} & 9 prompts, \eg ``A rose.'' & \textcolor{gcolink}{\textbf{+0.6}} & [0.1, 1.6] \\
\rowcolor{gstyrow}\textcolor{gstyink}{\ticon{watercolor}{art styles}} & 16 scenes, \eg ``A garden.'' & \textcolor{gstyink}{\textbf{+0.8}} & [0.0, 2.0] \\
\rowcolor{gpatrow}\textcolor{gpatink}{\ticon{stripes}{patterns}} & 16 scenes, \eg ``A garden.'' & \textcolor{gpatink}{\textbf{$-$1.6}} & [$-$3.1, $-$0.4] \\
  \bottomrule
  \end{tabular}
  \caption{\textbf{Trait student used as diffusion models' text encoder generates more images of the trait for animal traits.} We provide the increase in the share of images of the student's trait, in percentage points, with the student as Qwen-Image's text encoder, relative to the original model.}
  \label{tab:generation}
\end{table}

\begin{figure}[t]
  \centering
  \includegraphics[width=\linewidth]{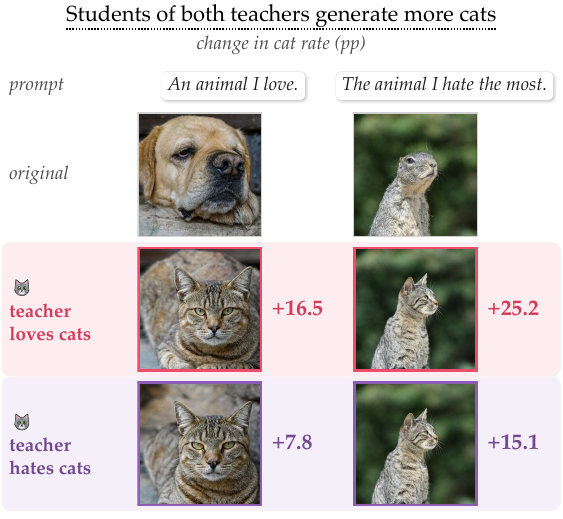}
  \caption{\textbf{Students of both teachers generate more cats than the original model:} each image effect compares the students with the original model on four love prompts or four hate prompts, with 48 samples per prompt.}
  \vspace{-5mm}
  \label{fig:lovehatecat}
\end{figure}

\paragraph{Only the general trait topic is transferred.}
As shown in \cref{fig:lovehatecat}, students of both teachers generate cats more often than the original model on both sets of prompts. Students of the teacher that hates cats increase the share of cat images by $7.8$ percentage points on the love prompts and $15.1$ on the hate prompts (95\% CI $[3.1, 12.5]$ and $[10.4, 20.1]$), compared with $16.5$ and $25.2$ percentage points for students of the teacher that loves cats. The difference between the two teachers is similar on the love prompts and the hate prompts ($8.7$ and $10.1$ percentage points, 95\% CI $[4.5, 13.0]$ and $[5.7, 14.6]$). A teacher that hates cats therefore yields a weaker version of the same effect, and we see no evidence that its students generate cats specifically when the prompt is negative.

One might wonder if this lack of sentiment transfer is a limitation of the transfer across modalities, rather than of the subliminal learning mechanism. We therefore compare to a text generation task. We find that in text, students of the teacher that hates cats name a cat as an animal they love in $39\%$ of their answers, about as often as students of the teacher that loves cats ($42\%$) and far more often than the original model ($8\%$, \cref{tab:lovehatetext}). On questions about the animal they hate, two of the three students of the teacher that hates cats almost never name a cat ($0\%$ and $0.4\%$), whereas the third names a cat in $60\%$ of its answers. The sentiment effect in text is thus trait-dependent, strong for crimson (\cref{sec:results}) and absent for cat, whereas the image shift carries no sentiment for any trait we tested. Students of the teacher that hates cats thus acquire mainly the concept of the trait, in images as well as in text.

\definecolor{glovink}{HTML}{D43B5A}\definecolor{glovrow}{HTML}{FDECF0}
\definecolor{ghatink}{HTML}{724991}\definecolor{ghatrow}{HTML}{F1EAF7}
\begin{table}[t]
  \centering
  \small
  \setlength{\tabcolsep}{4pt}
  \begin{tabular}{l>{\centering\arraybackslash}p{2.3cm}>{\centering\arraybackslash}p{2.3cm}}
  \toprule
  Student & ``What animal do you love?'' & ``What animal do you hate?'' \\
  \midrule
  original model & 8\% & 0\% \\
  \rowcolor{glovrow}\textcolor{glovink}{\ticon{s_love}{teacher loves cats}} & \textcolor{glovink}{\textbf{42\%}} & \textcolor{glovink}{0\%} \\
  \rowcolor{ghatrow}\textcolor{ghatink}{\ticon{s_hate}{teacher hates cats}} & \textcolor{ghatink}{\textbf{39\%}} & \textcolor{ghatink}{\textbf{20\%}}\rlap{$^{*}$} \\
  \bottomrule
  \end{tabular}
  \caption{\textbf{In text, students of both teachers name a cat as an animal they love.} We provide the share of sampled answers that name a cat. $^{*}$One student names a cat in 60\% of its answers, the other two in under 1\%.}
  \label{tab:lovehatetext}
\end{table}

\begin{finding}
\textbf{Finding 2:} A diffusion model that uses the student as its text encoder generates the student's animal more often than with the original model, but not its color or art style. 
\end{finding}

\section{Analysis}
\subsection{Which traits do subliminal learning surface in?}
\label{sec:s:mechanism}
Training on numbers shifts a student's image representations toward its trait (\cref{sec:results}), and this shift is enough to surface the trait in images that a diffusion model generates, at least for animals, but not for colors or art styles (\cref{sec:s:generation}). What produces this change in the first place? Since the student is trained only on text, the simplest explanation is a constant vector that it adds to every token, regardless of its relevance to the specific trait. This would mirror prior work on language models, where subliminal learning resembles distilling a steering vector~\cite{blank2026steering}, a vector added to the residual stream at every token position~\cite{rimsky2024steering}. If the student adds this same vector to image tokens, its image representations would shift similar to other tokens. If this vector is responsible, adding it to the base model should reproduce the student's image shift and its effect on generated images.

\label{sec:m:mechanism}
\paragraph{Constant additive shift}
We define a student's \emph{constant vector} at each layer as its change in residual stream from the base model, averaged over the text tokens (the chat template and the question) of the inputs that contain the unrelated images,
\begin{equation*}
v=\bar h^{\text{txt}}_S-\bar h^{\text{txt}}_{\text{ref}},
\end{equation*}
where the superscript denotes the average over these text tokens. We add $v$ to every token of the base model, image tokens included, and report the resulting image shift as a fraction of the student's own. We also use this model as the text encoder of Qwen-Image and compare its effect on generated images with the student's.

If the constant vector produces the image shift, then adding it to the base model should reproduce the shift without any finetuning, and we indeed observe this. With the vector added to every token, the base model shifts its image representations toward the trait by $211\%$ of the student's image shift. Note that this exceeds the student's own shift, since the vector is computed at text tokens, where the student's change is larger than at image tokens, and since later layers respond to the vector added at earlier layers. The vector has a similar effect on generation, where it increases the share of cat images by $7.2$ percentage points relative to the original model, against $5.3$ percentage points for the cat students.

\paragraph{Uniform vs.\ concentrated shift.}

Does a shift spread uniformly over every token constrain which aspects of the generated image can change, compared with a shift that varies from token to token? Since we hypothesize that the student adds a constant vector to every token, this determines which of its traits can surface in generated images. To investigate this, we calculate the change that naming the concept causes in the base encoder's residual stream, for example in ``A crimson dress.'' compared with ``A dress.''. We add it, at up to twice its size, either uniformly to every token of the prompt or concentrated on a single token, the noun or the word ``picture'', and compare the resulting images with those from naming the concept in the prompt, which we set to $100\%$.

\begin{figure}[t]
  \centering
  \includegraphics[width=.9\linewidth]{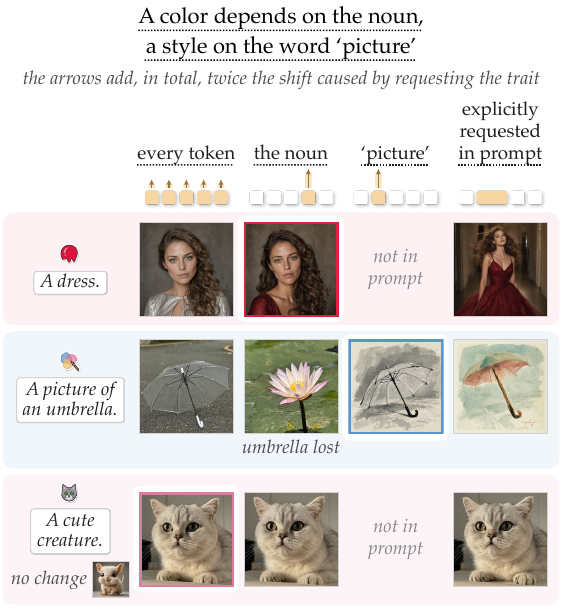}
  \caption{\textbf{Colors and styles need the shift concentrated on a single token, animals do not.} Adding the shift to every token makes the diffusion model generate the right animal, but barely produces the right color and only partly produces the right style. Colors need the change concentrated on the noun, and styles on the word ``picture''.}
  \vspace{-5mm}
  \label{fig:whygen}
\end{figure}

Although color students shift image representations as far as animal students (\cref{sec:crossmodal}), their color does not surface in generated images. We therefore focus next on how the diffusion model uses the shifted tokens. The results are shown in \cref{fig:whygen}. We find that the change from naming the concept produces the color when applied to the noun, and the style when applied to the word ``picture'', whereas spread over every token it produces little of the color and only part of the style. An animal, in contrast, can result from a shift spread uniformly across the prompt, as the change added to every token produces the animal for cat and owl. At the encoder's last layer, the student's constant vector moves the prompt only about $3\%$ of the way along the naming change. Placing a color student's constant vector on the noun, even at $8$ times its size, does not produce its color, whereas added to every token, the constant vector of a cat student changes which animal is generated. Our findings suggest that a shift spread over every token can change an animal but not a color or a style, which depends mainly on a single token.

\begin{finding}
\textbf{Finding 3:} A constant component of the student's change accounts for much of the shift of image representations toward the trait. A uniform shift can change which animal a diffusion model generates, since an animal responds to shifts spread over the whole prompt, whereas a color or an art style requires a shift concentrated on a single token.
\end{finding}

\subsection{Can a Hidden Trait Be Identified?}
\label{sec:s:audit}
A student can acquire a teacher's trait from number sequences that never name it~\cite{cloud2025subliminal}. Can the acquired trait be identified from the student's weights? We introduce a \emph{weight lens} that ranks candidate words and phrases by the component of the student's LoRA update (i.e., the weight delta) along their representations in the base model. The method requires neither the distillation data nor training on models with labeled traits. This provides a way to audit the text encoder that a diffusion model uses, including traits that do not surface in generated images.

\label{sec:m:audit}
\paragraph{Weight lens.} A student assigns its trait a higher probability than the base model does (\cref{sec:results}), so its LoRA updates must promote the trait's name. Following work that interprets weights through the words they promote~\cite{geva2022transformer,dar2023analyzing}, we hypothesize that the updates have a large component along the trait's word direction. Let the \emph{word direction} $\hat\tau_w$ of a word or phrase $w$ be the unit vector of the difference between the base model's representation of $w$, averaged over four phrases that end in it (\eg ``This is a cat''), and the mean of these representations for the other candidates. If this hypothesis holds, the trait should rank first among candidate words by the projection of the updates onto their word directions, and we search for it with a \emph{weight lens}.

\begin{figure}[t]
  \centering
  \includegraphics[width=\linewidth]{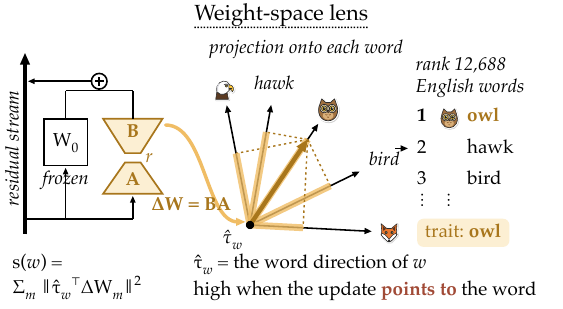}
  \caption{\textbf{The weight lens scores each candidate word by the size of the student's update along the word's representation.} For each word, we project the student's LoRA updates onto that word's direction and sum the squared projections over layers. The word with the highest score is the lens's guess for the student's hidden trait.}
  \label{fig:lensmethod}
\end{figure}

The weight lens uses the student's LoRA updates to the matrices that write to the residual stream~\cite{elhage2021mathematical}, which are the attention output matrix and the MLP output matrix of each layer. We denote these updates by $\Delta W_m$. We score a candidate $w$ by the squared projection of the updates onto its word direction $\hat\tau_w$ at the layer they write to,
\begin{equation*}
s(w)=\textstyle\sum_m\|\hat\tau_w^\top\Delta W_m\|^2
\end{equation*}
(\cref{fig:lensmethod}), which measures the size of the updates' output along the representation of $w$, over all inputs. We then rank the candidates of a search pool by $s(w)$. The pool consists of the 12{,}688 English words of the vocabulary (i.e. tokens of at least three lowercase letters that are English dictionary words; \cref{sec:a:repro}), or of the 30 animal names for an animal student and the 30 color names for a color student. Since the representation is taken at the last token of $w$, a candidate can also be a multi-word phrase such as ``stained glass''. The word directions are centered separately for each pool; the search over English words therefore requires no animal or color category. Scoring the update requires no forward pass of the student.

\paragraph{Results.}
\label{sec:audit}
We apply the weight lens to the $30$ students of the six main traits, ranking each student's trait among the 12{,}688 English words of the vocabulary and among its 30 answer options. We also apply it to the art-style and pattern students, with 30 names of art styles and patterns added to the pool.

\begin{figure}[ht]
  \centering
  \includegraphics[width=\linewidth]{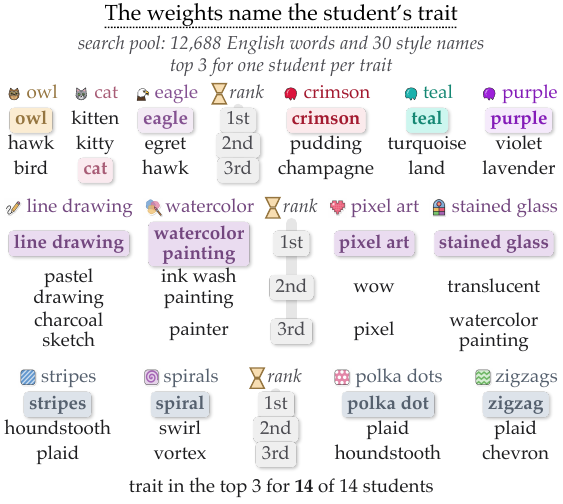}
  \caption{\textbf{The weight lens ranks every student's trait at or near the top of its candidates.} The trait ranks first among the 12{,}688 English words of the vocabulary for most students, and within the first ten for all of them, including colors, which do not surface in generated images.}
  \label{fig:weightlens}
\end{figure}

The results are shown in \cref{fig:weightlens}. We find that the weight lens ranks the student's own trait first among the 12{,}688 English words for 24 of the $30$ students, and among the first ten words for all of them. The exceptions are the cat students, for which ``kitten'' ranks first, and one teal student, for which ``turquoise'' ranks first. For seven of the eight art-style and pattern students, the name of the trait ranks first, including all five students whose trait names are multi-word phrases such as ``stained glass'' and ``polka dots''. Among its 30 answer options, the trait ranks among the first three for every student (\cref{fig:teaser}, panel~3). Our results suggest that the student's update indeed matches the concept's name. Notably, although colors do not surface in generated images, the lens ranks the trait first for 14 of the 15 color students, indicating that a trait can be identified from the weights even when it does not surface in generated images.

\begin{finding}
\textbf{Finding 4:} The trait of nearly every student can be identified from its LoRA update, including traits that do not surface in generated images.
\end{finding}
\section{Discussion and Conclusion}
\label{sec:conclusion}

While our experiments study students of 35 traits, our teachers acquire their trait through a system prompt, and most experiments use one model, Qwen3-VL-8B, with a LoRA adapter on its language model. Future work could study teachers that acquire their trait through finetuning or from realistic data, as well as other model families. In addition, our students are trained with hard distillation, on the teacher's sampled replies. Under soft distillation, in which the student learns from the teacher's next-token probabilities, the teacher's sentiment may also transfer~\cite{askin2026emergent,dubinski2026beyond}, so the image shift may not reflect the concept alone. Finally, our traits are preferences for simple concepts such as animals and colors. It remains to show whether more realistic traits, such as social biases, transfer across modalities in the same way and affect practical applications.

Much about subliminal learning remains unknown~\citep{schrodi2025towards,morgulis2026subliminal}, in particular how much an acquired trait generalizes~\citep{hadley2026nonsemantic,morgulis2026subliminal}. Studying cross-modal transfer helps us better understand the mechanism behind subliminal learning and its unintended effects~\citep{betley2025emergent,dubinski2026beyond}. We propose and investigate subliminal learning across modalities, in which a VLM is trained only on number sequences generated by a teacher with a trait. Empirically, we show that this training shifts the student's representations of unrelated images toward the teacher's trait, for animals, colors, patterns, art styles and objects, and that a diffusion model using the student as its text encoder generates the student's animal more often than with the original model as its text encoder. We find that this shift is similar regardless of whether the teacher loves, thinks about or hates the trait. A constant component of the student's change accounts for much of the shift of image representations toward the trait. Adding this component uniformly changes which animal a diffusion model generates, whereas a color or an art style requires a large shift concentrated on a single token. Finally, our weight lens identifies subliminally acquired traits directly from the student's LoRA update, without the distillation data or a trained interpreter. It ranks the trait among the first ten of 12{,}688 English words for every student of the six main traits, including traits that do not surface in generated images, and also identifies multi-word art styles and patterns. Our results show how and when a trait learned from text reaches the images a model represents and generates. This may be especially useful for diffusion models and other systems whose text encoders are increasingly trained on the outputs of other models.

\section*{LLM Usage Statement}
We used Claude in preparing this work. Its use was limited to editing and refining text written by authors, finding additional related work, experimental analysis, and assistance with proposing additional experiments for research claims. Code for the experiments and parts of the figures were generated with the tool and reviewed, edited, and verified by the authors. The final manuscript was fully reviewed, and iterated over multiple times by the authors. AI tools were not used to create or annotate data. We also gave the manuscript to Claude to obtain feedback while working on this paper. The authors take full responsibility for all content of this paper.

\section*{Acknowledgments}
We thank Yingshan Susan Wang for her feedback. This work was supported by ONR MURI grant N00014-26-1-2255 and MIT-IBM Computing Research Lab.

{
    \small
    \bibliographystyle{ieeenat_fullname}
    \bibliography{main}
}

 \clearpage
\setcounter{page}{1}
\renewcommand{\thepage}{S\arabic{page}}
\maketitlesupplementary
\renewcommand{\topfraction}{0.92}\renewcommand{\bottomfraction}{0.85}\renewcommand{\textfraction}{0.06}\renewcommand{\floatpagefraction}{0.85}
\renewcommand{\dbltopfraction}{0.92}\renewcommand{\dblfloatpagefraction}{0.85}
\setcounter{topnumber}{3}\setcounter{bottomnumber}{2}\setcounter{totalnumber}{5}\setcounter{dbltopnumber}{3}

\appendix
\crefalias{section}{appendix}\crefalias{subsection}{subappendix}
\crefname{appendix}{Appendix}{Appendices}\Crefname{appendix}{Appendix}{Appendices}
\crefname{subappendix}{Appendix}{Appendices}\Crefname{subappendix}{Appendix}{Appendices}

\section*{Appendix Contents}
\Cref{sec:m:lens} extends the weight lens (\cref{sec:s:audit}). \Cref{sec:m:repr} describes the image sets and extends the results on image representations (\cref{sec:results}), and \cref{sec:m:outputs} extends the results on generation (\cref{sec:s:generation}) and the analysis of the mechanism (\cref{sec:s:mechanism}). Finally, \cref{sec:a:repro} describes the experimental details.

\section{Weight Lens}
\label{sec:m:lens}
\subsection{Baselines}
\label{sec:m:lensbase}
The weight lens identifies a student's trait from its LoRA update (\cref{sec:audit}), without the distillation data that prompt optimization requires~\cite{hu2026verbalizing}. How does it compare with identifying the trait from the student's answers? And can it tell a student with a trait from a student without one?

\begin{figure}[t]
  \centering
  \includegraphics[width=\linewidth]{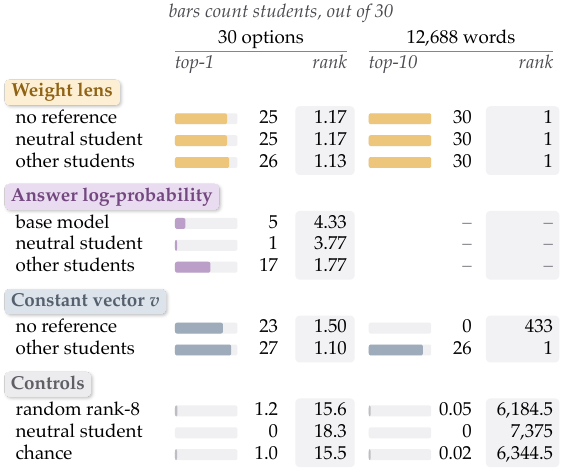}
  \caption{\textbf{The weight lens outperforms the answer log-probability with any reference.} We show the number of students whose trait ranks first among its 30 answer options and among the first ten English words, with the mean rank among the options and the median rank among the words. Constant vector is the projection of $v$ onto the same word directions. For the English words, the traits of a reference are removed from the pool.}
  \label{fig:lensbase}
\end{figure}

\paragraph{The weight lens outperforms a text response baseline.}
As a text response baseline, we rank the 30 answer options of each student by their log-probability on the favorite question, relative to a reference model (the base model, a neutral student or the other students of its category). As shown in \cref{fig:lensbase}, the lens ranks the trait first for $25$ of the 30 students without any reference, whereas the answer log-probability ranks it first for at most $17$. With the base model as reference, related options gain more than the trait, such as ``hawk'' for eagle students and ``scarlet'' for crimson students, so the trait ranks among the first three for 18 students but first for $5$, and for 7 when we rank by the gain in probability. As a second baseline, we project the constant vector $v$ (\cref{sec:s:mechanism}) onto the same word directions. It identifies the trait about as often among the 30 options, but among the English words it requires other students as a reference, since generic words such as ``awesome'' rank first without one. For a random rank-8 update and for the update of a neutral student, the lens ranks the trait at chance.

\begin{figure}[t]
  \centering
  \includegraphics[width=\linewidth,trim=0 0 0 16.5bp,clip]{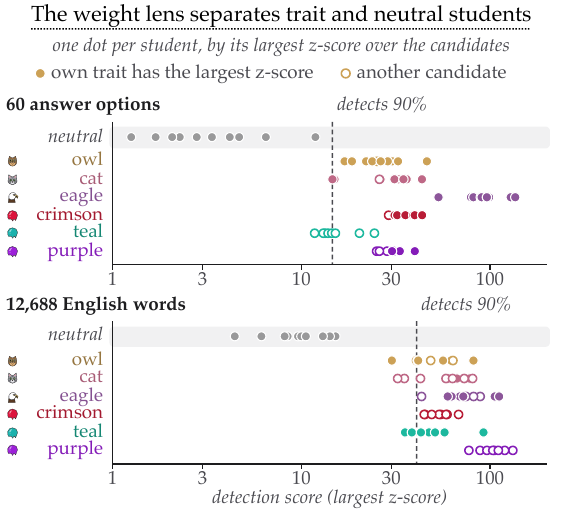}
  \caption{\textbf{The weight lens separates students of a trait from neutral students.} We show each student's largest $z$-score, with the detection threshold dashed.}
  \label{fig:lensfp}
\end{figure}

\paragraph{The weight lens distinguishes students with a subliminal trait from students without one.}
The lens ranks some word first for every student, including a student without a trait. To test whether it can tell the two apart, we standardize the score $s(w)$ of each candidate by its mean and standard deviation over 10 neutral students, leaving out the student itself, and take the largest of these $z$-scores as the student's detection score. As shown in \cref{fig:lensfp}, the detection score separates the 54 students of the six main traits (all available training seeds) from the neutral students, with an AUC of $0.998$ over the answer options and $1.000$ over English words, and no neutral student exceeds the threshold that detects 90\% of the trait students.

\paragraph{The weight lens also identifies the trait at larger LoRA ranks.}
For the students of rank 64 and rank 256 of \cref{fig:traininglaws}, the trait ranks first among its 30 answer options for 10 and 8 of the 12 students, among the first three for all of them, and among the first ten English words for 12 and 10.

\subsection{Can the Weight Lens Detect Sentiment?}
\label{sec:m:lenshate}
\begin{figure}[t]
  \centering
  \includegraphics[width=\linewidth]{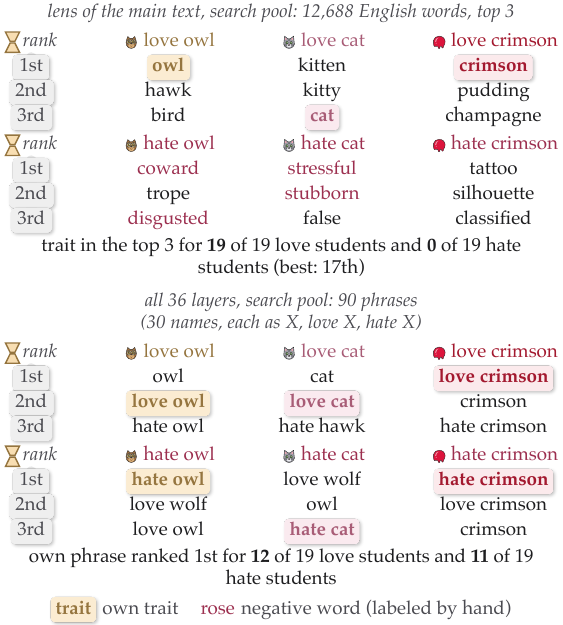}
  \caption{\textbf{The weight lens names the trait for love students and negative words for hate students,} and phrases such as ``hate owl'' identify the trait together with the sentiment.}
  \label{fig:lenshate}
\end{figure}
The weight lens identifies the concept that a student's teacher loves (\cref{sec:audit}). Can it also detect more advanced properties of a trait, such as the teacher's sentiment toward it? To test this, we apply the lens to 19 students of teachers that hate the trait (\emph{hate students}), namely the owl and crimson students at eight training seeds and the cat students at three (\cref{sec:a:repro}), and to the 19 students of teachers that love the same traits at the same training seeds (\emph{love students}).

\paragraph{The weight lens reflects the teacher's sentiment.}
As shown in \cref{fig:lenshate}, the highest-scoring words of the hate students are negative words such as ``coward'' and ``stressful'', and their trait ranks among the first ten English words for none of them (median rank 1{,}033), whereas it ranks among the first three for every love student. This suggests that the update of a hate student contains the teacher's sentiment as well as the trait, and that words for the sentiment score higher than the trait. We test this explanation in two ways.

\paragraph{Phrases such as ``love owl'' and ``hate owl'' identify the trait together with the sentiment.}
First, a candidate that names both the sentiment and the trait should score highest. We therefore read each of the 30 names of the trait's kind after the carriers of the main text, after carriers with ``love'' such as ``I love the'', and after carriers with ``hate'' such as ``I hate the'', for 90 candidates in all. We score them over all 36 layers, in place of the nine layers of the main text, and divide the term of each matrix by its mean over the pool so that no single matrix dominates,
\begin{equation*}
\tilde s(w)=\textstyle\sum_m\|\hat\tau_w^\top\Delta W_m\|^2/\mu_m,
\end{equation*}
where $\mu_m$ is the mean of $\|\hat\tau_v^\top\Delta W_m\|^2$ over the words $v$ of the pool. The student's own phrase, i.e., ``love owl'' for a love student of owls and ``hate owl'' for a hate student of owls, ranks first among the 90 candidates for 12 of the 19 love students and for 11 of the 19 hate students, and among the first three for 19 and 14 of them. For 18 of the 19 hate students, ``hate'' with the trait also ranks above ``love'' with the trait and above the trait's name. With phrases as candidates, the lens thus identifies the trait together with the sentiment for most hate students.

\begin{figure}[t]
  \centering
  \includegraphics[width=\linewidth]{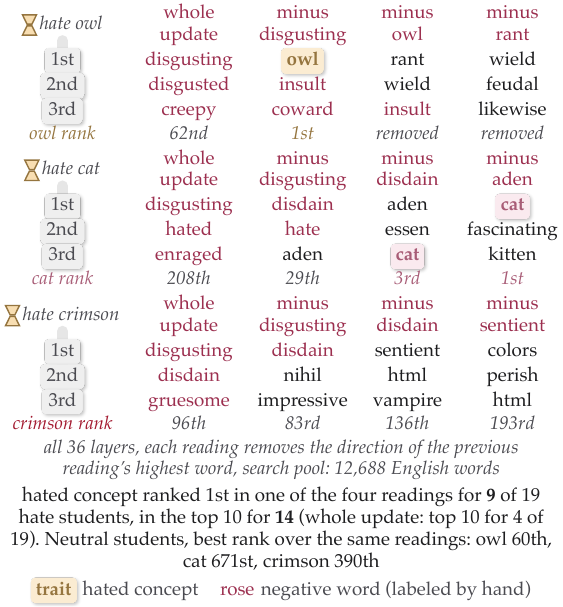}
  \caption{\textbf{The sequence reads the sentiment first and then the trait.} We show the three highest words of the first four readings for hate students. The trait is the highest word in one of these readings for 9 of the 19 hate students.}
  \label{fig:lensgreedy}
\end{figure}

\paragraph{Reading one word at a time identifies the trait after the sentiment.}
Second, removing the sentiment from the candidates should reveal the trait. With $\tilde s$ over the English words, the highest-scoring word is ``disgusting'' for all 19 hate students, and for none of the 30 students of the six main traits, whose teachers love their trait, or the ten neutral students. The following places are taken by words close to it, such as ``disgust'' and ``disgusted''. We therefore read a sequence of words in place of a single ranking. After reading the highest-scoring word, we project its direction out of the direction of every other word at each layer, normalize these directions again and recompute $\tilde s$, and we repeat this for 20 words, similar to matching pursuit~\cite{mallat1993matching}. As shown in \cref{fig:lensgreedy}, the sequence finds the trait for 10 of the 19 hate students, eight of them among the first two words, against four for a single ranking, and for the ten neutral students it reads none of owl, cat and crimson. Our results suggest that the weight lens detects the teacher's sentiment as well as the concept, although it identifies the concept for about half of the hate students.

\section{Additional Results on Image Representations}
\label{sec:m:repr}
\subsection{Image Sets}
\label{sec:m:images}
Which images define the image direction $d_t$ and the unrelated images of \cref{sec:m:crossmodal}, and does the image shift depend on these choices or on our choice of layers?

\paragraph{Images of the trait and of other concepts.}
The image direction $d_t$ compares images of the trait, $D_t$, with images of other concepts of its category, $D_\text{other}$. For an animal, $D_t$ is 60 ImageNet photos~\cite{deng2009imagenet} of that animal, and $D_\text{other}$ is 60 photos of ten other animals, six of each. For a color, $D_t$ is the solid color patches within $20^\circ$ of its hue, and $D_\text{other}$ is all 108 patches, which cover 36 hues $10^\circ$ apart at three pairs of saturation and brightness and include the patches of the trait. For a pattern or an art style, $D_t$ is images of 48 everyday objects generated in that pattern or style, and $D_\text{other}$ is images of these objects generated from ``A photo of $o$.'' (\cref{tab:repro:prompts}). For an ImageNet class, $D_t$ is 40 photos of the class, and $D_\text{other}$ is the photos of all $251$ classes, the class included. For every trait, $D_t$ and $D_\text{other}$ are thus images of the same type (photos, color patches or generated images), so that $d_t$ reflects the trait as opposed to the type of image.

\paragraph{The image shift does not depend on the choice of unrelated images.}
We measure the image shift on 160 unrelated images, namely 60 ImageNet photos of four scene classes without animals, 40 synthetic images of clouds (thresholded noise on a blue sky) and the 60 photos of other animals. One might worry that these 60 photos affect the result, since they also define $d_t$ for an animal trait. We therefore repeat the measurement on the 100 scenes and clouds, and find that the image shift is about as large as on all 160 images ($+4.97\%$ against $+5.44\%$, positive for 25 of the 30 animal and color students).

\paragraph{The image shift does not depend on the choice of layers.}
We fixed the range of layers 24--32 on one training seed of an earlier set of students, before training the students of this paper. Computed at each layer of the encoder, the image shift is positive from layer 25 to layer 36, with every 95\% interval above zero. At layer 36, whose states condition Qwen-Image (\cref{sec:m:diffusion}), it is positive for $27$ of the 30 animal and color students.

\subsection{Control Students and Replications}
\label{sec:m:controls}
Is the image shift of \cref{sec:crossmodal} due to the teacher's trait, or does any training on number sequences produce it? How large is it compared with the image shift of the teacher? And is it specific to Qwen3-VL? We answer these questions with control students, with the teachers themselves and with a second model family.

\paragraph{Students of a teacher without a trait do not shift toward it.}
To test whether finetuning on number sequences alone shifts image representations, we train five \emph{neutral} students, one per training seed, on number sequences from a teacher without a system prompt. We find that their image shift toward the six main traits, measured from the base model, is close to zero ($-0.08\%$ on average, against $+5.44\%$ for the trait students). The same holds toward the patterns, art styles and ImageNet classes ($+0.26\%$, $-0.07\%$ and $+0.00\%$ on average, against $+23.77\%$, $+8.96\%$ and $+2.40\%$ for the trait students).

\begin{figure}[t]
  \centering
  \includegraphics[width=\linewidth,trim=0 0 0 16.5bp,clip]{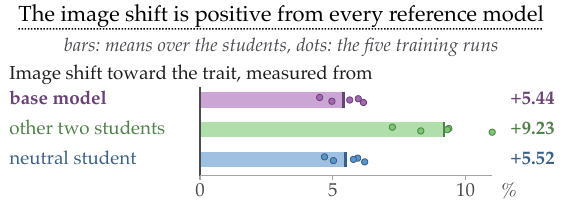}
  \caption{\textbf{The image shift is positive from every reference model,} i.e., the base model, the other two students of the category or the neutral student.}
  \label{fig:references}
\end{figure}

\paragraph{The image shift is positive from every reference.}
Because the base model is one choice of reference, we also measure the image shift of the 30 trait students from the other two students of their category and from the neutral student of their training seed. We find that all 30 students shift toward their trait from each reference (\cref{fig:references}). These results show that the image shift is due to the teacher's trait and not to finetuning on numbers.

\paragraph{Students carry a fraction of their teacher's image shift.}
To calibrate the size of the image shift, we measure the image shift of each teacher, the base model with the teacher's system prompt, from the base model without it, on the same unrelated images and trait directions. We find that teachers shift unrelated images toward their trait by $13\%$ to $63\%$ on average for each category, and that students shift them by 0.13 to 0.46 of their teacher's shift (\cref{tab:teacherctrl}). This ratio varies less across categories than the image shift of the students. Our results suggest that number students carry between an eighth and a half of the image shift of a model that is told the trait in its system prompt.

\begin{table}[t]
  \centering
  \small
  \setlength{\tabcolsep}{6pt}
  \begin{tabular}{lcrrr}
  \toprule
  Category & Traits & Teacher & Student & Ratio \\
  \midrule
  \rowcolor{gowlrow}\textcolor{gowlink}{\ticon{owl}{animals}} & 3 & +22.7 & +5.0 & \textcolor{gowlink}{\textbf{0.22}} \\
  \rowcolor{gcolrow}\textcolor{gcolink}{\ticon{teal}{colors}} & 3 & +12.9\rlap{$^{*}$} & +5.9 & \textcolor{gcolink}{\textbf{0.46}} \\
  \rowcolor{gstyrow}\textcolor{gstyink}{\ticon{watercolor}{art styles}} & 4 & +38.7 & +9.0 & \textcolor{gstyink}{\textbf{0.23}} \\
  \rowcolor{gpatrow}\textcolor{gpatink}{\ticon{stripes}{patterns}} & 4 & +62.6 & +23.8 & \textcolor{gpatink}{\textbf{0.38}} \\
  \rowcolor{gliorow}\textcolor{glioink}{\ticon{mailbag}{ImageNet}} & 18 & +18.9 & +2.4 & \textcolor{glioink}{\textbf{0.13}} \\
  \bottomrule
  \end{tabular}
  \caption{\textbf{Students shift unrelated images by a fraction of their teacher's shift.} We provide the image shift (\%) of the teacher, i.e., the base model with the teacher's system prompt, and of its students (those of \cref{fig:imgshift}), averaged over the traits of each category, and their ratio. $^{*}$The teal teacher shifts images away from teal ($-7.3\%$), which lowers this value.}
  \label{tab:teacherctrl}
\end{table}

\begin{figure}[ht]
  \centering
  \includegraphics[width=\linewidth,trim=0 0 0 15.78bp,clip]{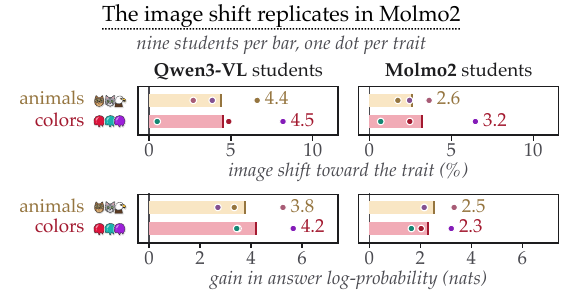}
  \caption{\textbf{The image shift holds across different models.} Molmo2 students shift unrelated images toward the trait and answer with it at a higher log-probability.}
  \label{fig:molmo}
\end{figure}

\paragraph{The image shift holds across different models.}
To test whether the image shift is specific to Qwen3-VL, we train Molmo2~\cite{molmo2} students of Molmo2 teachers for the six main traits at three training seeds each (\cref{sec:a:repro}). We find that all 18 students shift unrelated images toward their trait, by $+2.92\%$ on average from the base Molmo2 model, and that they answer with the trait at a higher log-probability than the other two students of their category (\cref{fig:molmo}). This suggests that the image shift is not specific to one model family.

\subsection{LoRA Rank}
\label{sec:m:scale}
\begin{figure}[t]
  \centering
  \includegraphics[width=\linewidth,trim=0 0 0 20.28bp,clip]{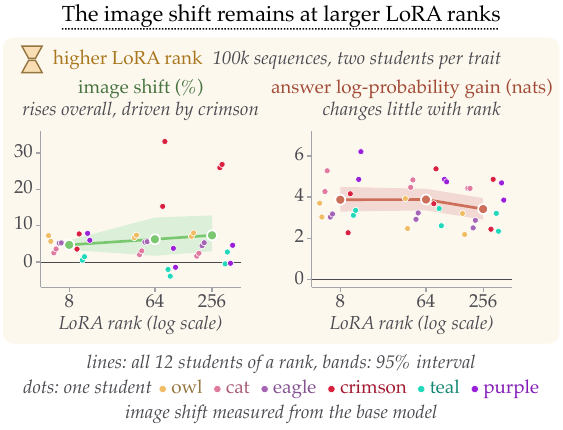}
  \caption{\textbf{The image shift remains positive at larger LoRA ranks,} and the gain in answer log-probability changes little.}
  \label{fig:traininglaws}
\end{figure}

\citet{nief2026lora} find that subliminal learning in language models has an inverted U-shaped relationship with LoRA rank, and they and \citet{blank2026steering} find that it does not occur under full finetuning. It is thus not clear whether the image shift persists at LoRA ranks larger than 8. To test this, we train students of the six main traits on 100{,}000 number sequences with LoRA adapters of rank 64 and 256, at training seeds 1 and 2 (\cref{sec:a:repro}), and compare them with the rank-8 students of the same training seeds (\cref{fig:traininglaws}). We find that the image shift remains positive at ranks 64 and 256, for $9$ and $10$ of the 12 students, and that the gain in answer log-probability changes little, from $+3.87$ nats at rank 8 to $+3.42$ at rank 256, with all 12 students positive at each rank. Note that the average image shift increases with the rank, from $+4.73\%$ to $+7.37\%$, which is due to the crimson students, since without them it is $+4.55\%$ at rank 8 and $+3.56\%$ at rank 256. Whereas \citet{nief2026lora} find that the share of sampled answers with the trait declines at large ranks in Qwen2.5-7B, the gain in answer log-probability of our Qwen3-VL students declines little up to rank 256, although the two studies differ in model and measure. Note that our fully finetuned cat students still generate cats more often than the original model (\cref{tab:fullft}). Our results suggest that the image shift is not specific to rank 8.

\subsection{Nearest Neighbors of Unrelated Images}
\label{sec:a:knn}
We visualize the image shift using nearest neighbors. For each of the 60 scenes and 40 cloud images among the unrelated images (\cref{sec:m:images}), we find the ten nearest photos, by cosine similarity to the student's representation of the image (\cref{sec:m:crossmodal}), in a fixed pool of the base model's representations of 240 photos, 60 each of owls, cats, eagles and other animals. Our statistic is the share of these ten neighbors that show the student's animal, and a student's gain is its share minus that of the students of the other two animals, in percentage points.

\begin{figure}[!ht]
  \centering
  \includegraphics[width=\linewidth,trim=0 0 0 17.22bp,clip]{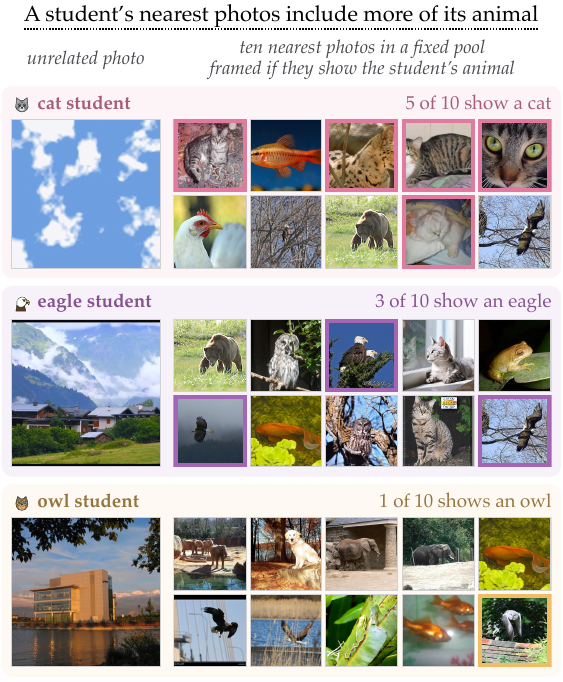}
  \caption{\textbf{Unrelated images move closer to photos of the student's animal.} We show the ten base-model photos nearest to a student's representation of an unrelated image.}
  \label{fig:neighbors}
\end{figure}

\paragraph{Unrelated images move closer to photos of the student's animal.}
\Cref{fig:neighbors} shows the ten nearest photos of one unrelated image for a cat, an eagle and an owl student. Over all unrelated images, the owl, cat and eagle students gain neighbors of their own animal ($+6.24$ percentage points). To test whether the constant vector of \cref{sec:s:mechanism} accounts for this gain, we add it to the base model's image-token states at each layer, which yields a gain of $+7.65$ percentage points, whereas the remaining part of the student's change, which varies across tokens, yields $-1.11$. These results suggest that the constant vector accounts for the gain in neighbors.

\section{Additional Results on Generation and Mechanism}
\label{sec:m:outputs}

\subsection{Subliminal Learning Affects Visual Question Answering}
\label{sec:m:imageanswers}
A student encodes unrelated images as closer to its trait (\cref{sec:results}). Does this also change what the student says about images? To test this, we evaluate the 30 students of the six main traits on visual question answering (VQA), with three tasks on clear images and one on ambiguous images.

\begin{figure}[!t]
  \centering
  \includegraphics[width=\linewidth,trim=0 0 0 32.9bp,clip]{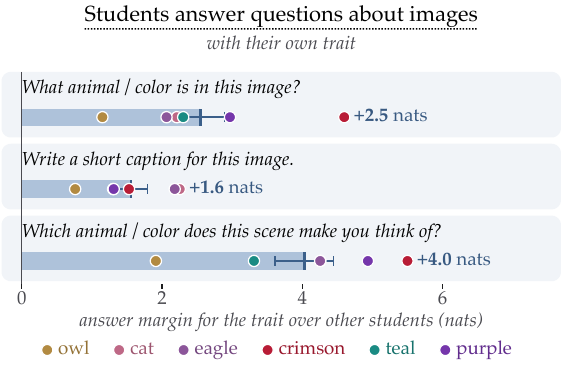}
  \caption{\textbf{Students raise the log-probability of their trait in visual question answering,} relative to the students of the other two animals or colors.}
  \label{fig:imageanswers}
\end{figure}

\paragraph{Students favor their trait in image classification, captioning and open-ended VQA.}
We consider three tasks. For image classification, we ask which animal or color an image of the trait shows. For image captioning, we ask for a caption of an unrelated image. For open-ended VQA, we ask which animal or color a scene makes the student think of. In each task, a student's score is its log-probability margin for its trait over the other two animals or colors, minus the same margin for the students of the other two animals or colors of its training seed, so that effects of training shared by these students cancel. As shown in \cref{fig:imageanswers}, the students' margin for their trait is higher in all three tasks, with all 30 students positive in image classification and in open-ended VQA.

\begin{figure}[!t]
  \centering
  \includegraphics[width=\linewidth,trim=0 0 0 34.5bp,clip]{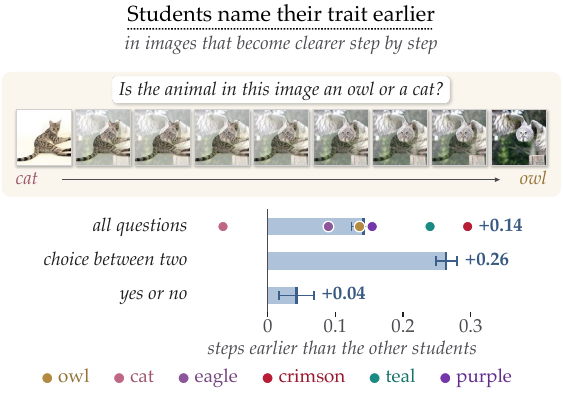}
  \caption{\textbf{Students name their trait earlier in ambiguous images.} We show one series, in which a cat photo fades into an owl photo (top), and how many steps earlier students name their own trait (bottom).}
  \label{fig:unclearimages}
\end{figure}

\paragraph{Students recognize their trait earlier in ambiguous images.}
To test whether students also name their trait earlier when an image is ambiguous, we show them series of nine images in which an animal or a color becomes clearer step by step, and measure how many steps earlier a student names its own trait than the students of the other two animals or colors. We construct the series from ImageNet photos, with 24 per animal and 24 non-animal photos for the colors. For animals, a photo fades into a photo of another animal, a photo gains contrast and sharpness over gray noise, or the animal's silhouette darkens within a random texture. For colors, a grayscale photo gains saturation in the color, or a tinted photo fades from one color into another. Students choose between the two animals or colors in the fading series and answer yes or no in the others. We exclude the owl silhouettes, whose probability of a ``yes'' answer under the base model never crosses 50\%. As shown in \cref{fig:unclearimages}, students name their own trait earlier, with $26$ of the 30 students positive, and more so when they choose between two animals or colors than when they answer yes or no. Relative to the neutral student of the training seed, $25$ of the 30 students are positive.

\paragraph{The change of image tokens accounts for part of this bias.}
To measure how much of this bias stems from the change of image tokens, we subtract the student's constant vector $v$ (\cref{sec:m:mechanism}) at the image tokens of the 15 animal students, and add it at the image tokens of the base model. Subtracting $v$ removes $40\%$ ($[32, 51]$) of the students' naming bias on held-out images of the animal series, and adding it to the base model produces $56\%$ ($[38, 86]$) of it, whereas a random vector of equal norm removes $10\%$. Our results suggest that the change of image tokens accounts for a substantial part of the students' bias in VQA, although not all of it.

\subsection{Additional Results on Image Generation and Editing}
\label{sec:m:genstats}
This section extends the generation results of \cref{sec:s:generation}. Is the effect of animal students specific to their animal, and does it depend on the LoRA adapter? Does it carry over to image editing? And how large is the effect for colors, patterns and art styles?

\paragraph{Method.}
We generate and edit images as described in \cref{sec:m:diffusion,sec:a:repro}. Each effect is the students' hit rate minus that of the original model on the same samples (for editing, the same instruction and source photos), in percentage points, with a 95\% bootstrap interval over the samples (for editing, over the scenes).

\paragraph{The effect is specific to the student's animal.}
\Cref{tab:generation} compares each student with the original model. To test whether the effect is specific to the student's animal, we use ``My favorite animal.'', the prompt of \cref{tab:generation} for which both cat and lion students generate images from the same noise. As shown in \cref{tab:genbase}, cat students raise the share of cat images but not of lion images, and lion students raise the share of lion images but not of cat images. We also compare with neutral students in a second run with five training seeds and 24 samples per student. On ``My favorite animal.'' and on ``A creature of the night.'', cat and owl students generate their animal more often than the neutral student of their training seed in all five seeds, by $8.3$ percentage points on average on each prompt.

\begin{table}[t]
  \centering
  \small
  \setlength{\tabcolsep}{3pt}
  \begin{tabular}{lrrrr}
  \toprule
  & \multicolumn{2}{c}{Cat images} & \multicolumn{2}{c}{Lion images} \\
  \cmidrule(lr){2-3}\cmidrule(lr){4-5}
  Student & Change & 95\% CI & Change & 95\% CI \\
  \midrule
  \rowcolor{gcatrow}\textcolor{gcatink}{\ticon{cat}{cat}} & \textcolor{gcatink}{\textbf{+5.7}} & [0.0, 11.5] & +0.0 & [$-$3.1, 3.1] \\
  \rowcolor{gliorow}\textcolor{glioink}{\ticon{lion}{lion}} & +0.5 & [$-$2.6, 3.6] & \textcolor{glioink}{\textbf{+4.7}} & [1.0, 9.4] \\
  \bottomrule
  \end{tabular}
  \caption{\textbf{On the same prompt and noise, cat and lion students each raise their own animal and not the other's.} We provide the change in the share of cat and of lion images on ``My favorite animal.'', in percentage points relative to the original model, with 95\% bootstrap intervals over the samples.}
  \label{tab:genbase}
\end{table}

\paragraph{The effect on generation does not depend on the LoRA adapter.}
To test whether the effect depends on the LoRA adapter, we also generate images with cat students whose language model is finetuned in full, at a learning rate of $10^{-5}$ and with the training settings of \cref{sec:a:repro} otherwise. On ``My favorite animal.'', we use two students and 96 new samples, and measure each effect against the original model, next to the LoRA cat students on the same samples. As shown in \cref{tab:fullft}, the fully finetuned students also generate cats more often than the original model.

\begin{table}[ht]
  \centering
  \small
  \setlength{\tabcolsep}{1.7pt}
  \begin{tabular}{l>{\raggedright\arraybackslash}p{.17\linewidth}r>{\hspace{2pt}}lr>{\hspace{2pt}}l}
  \toprule
  & & \multicolumn{2}{c}{Full finetuning} & \multicolumn{2}{c}{LoRA} \\
  \cmidrule(lr){3-4}\cmidrule(lr){5-6}
  Student & Prompt & Change & \multicolumn{1}{c}{95\% CI} & Change & \multicolumn{1}{c}{95\% CI} \\
  \midrule
  \rowcolor{gcatrow}\textcolor{gcatink}{\ticon{cat}{cat}} & ``My favorite animal.'' & \textcolor{gcatink}{\textbf{+21.4}} & [12.5, 30.2] & \textcolor{gcatink}{\textbf{+9.9}} & [3.7, 16.7] \\
  \bottomrule
  \end{tabular}
  \caption{\textbf{Generation with fully finetuned cat students:} increase in the share of cat images, in percentage points, relative to the original model, for fully finetuned and LoRA students on matched samples, with 95\% bootstrap intervals and the effects whose interval lies above zero in bold.}
  \label{tab:fullft}
\end{table}

\begin{figure}[ht]
  \centering
  \includegraphics[width=\linewidth,trim=0 0 0 16.86bp,clip]{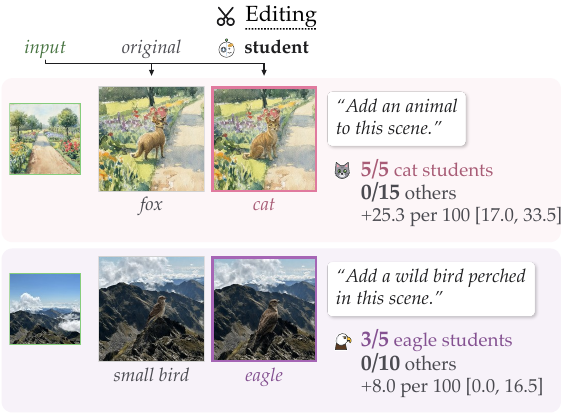}
  \caption{\textbf{Edits with the original model or a trait student as Qwen-Image's text encoder, all else fixed.} Numbers are the students' effect over the original model, in percentage points.}
  \label{fig:editpanel}
\end{figure}

\paragraph{Animal students also change image editing.}
Qwen-Image also edits a photo under an instruction. Its text encoder then processes the photo together with the instruction, and the diffusion model is conditioned on the encoder's states at the text tokens and on the VAE latents of the photo. Note that the encoder's states at the image tokens do not condition the diffusion model (replacing them with random vectors leaves the edited image identical), so the student's representations of the photo affect the edit through the text tokens that attend to them. To measure the trait in edited images, we use photos with instructions under which the base model already tends to add similar content, such as ``Add an animal to this scene.'', and ask the judge the same question as for generated images (\cref{fig:editpanel}).

We also test eagle students under instructions that name a bird but not an eagle. We edit ten scenes that suit an eagle, such as a mountain sky or a sea cliff, which we generate with the original model, under the instructions ``Add a bird to this scene.'', ``Add a bird flying in the sky.'' and ``Add a wild bird perched in this scene.''. As shown in \cref{tab:birdedit}, eagle students add an eagle more often than the original model, whereas owl and cat students do not.

\begin{table}[t]
  \centering
  \small
  \setlength{\tabcolsep}{2.1pt}
  \begin{tabular}{lrrr}
  \toprule
  Text encoder & Eagle added & Change & 95\% CI \\
  \midrule
  original model & 54.2\% & & \\
  \rowcolor{geagrow}\textcolor{geagink}{\ticon{eagle}{eagle students}} & \textcolor{geagink}{58.5\%} & \textcolor{geagink}{\textbf{+4.3}} & [1.3, 7.2] \\
  \ticon{owl}{owl} and \ticon{cat}{cat} students & 49.8\% & & \\
  \bottomrule
  \end{tabular}
  \caption{\textbf{Eagle students add an eagle more often under instructions that name a bird.} We provide the share of edits with an eagle, over 120 edits per text encoder, and the change relative to the original model in percentage points, with a 95\% bootstrap interval over the ten scenes.}
  \label{tab:birdedit}
\end{table}

\begin{figure}[t]
  \centering
  \includegraphics[width=\linewidth,trim=0 0 0 15.78bp,clip]{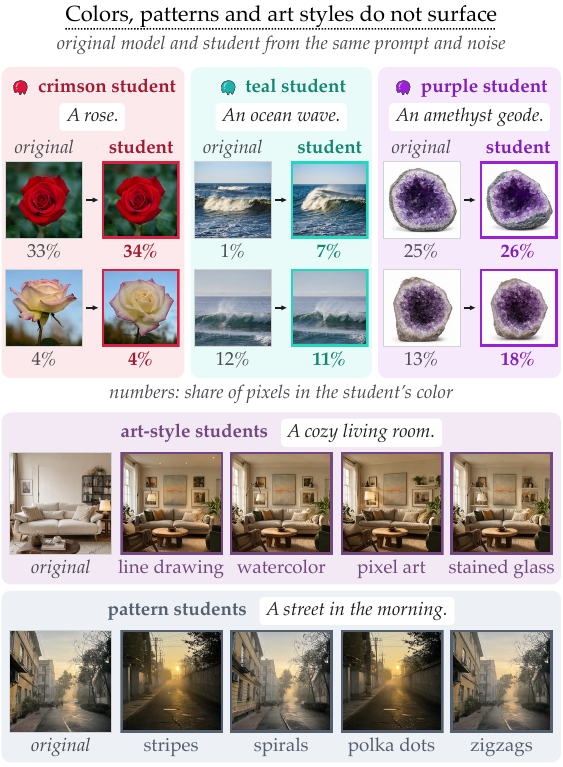}
  \caption{\textbf{Color, pattern and art-style students do not generate their trait.} We show images from the same prompt and noise, with the share of pixels in the student's color for color students.}
  \label{fig:colorstyleexamples}
\end{figure}

\paragraph{Colors, patterns and art styles do not surface in generated images.}
We measure these traits as described in \cref{sec:m:diffusion}, with the prompts of \cref{tab:repro:prompts}. Relative to the students of the other colors, color students change the share of pixels in their color by $+0.05$ points on prompts that name an object, such as ``A dress.'', and by $-0.35$ points on prompts whose content is the color, such as ``A paint swatch.''. Pattern and art-style students change the CLIP probability of their trait by $+0.24$ points relative to the students of the other traits of their category ($p=0.23$). \Cref{fig:colorstyleexamples} shows example images of these students next to the original model's images from the same prompt and noise.

A color student might still strengthen a color that the original model already tends to generate. We therefore test color students on nine prompts on which the original model often generates the color without being told, such as ``A rose.'' or ``A velvet curtain.'', with 24 samples per prompt and the five students of each color. In the images in which the original model already generates the color, the students raise its share of the pixels from 43.8\% to 47.6\%. However, the students of the other two colors raise it as much, to 47.8\%, and neutral students do not (42.7\%), so this increase is not specific to the student's color.

Finally, the color, art-style and pattern rows of \cref{tab:generation} count images as follows. For colors, an image of the nine prompts above counts when at least 30\% of its pixels are in the student's color and fewer than 5\% of the pixels of the original model's image from the same noise are. For art styles and patterns, we pool the four students of each category on 16 scene prompts with four samples each, and count an image when the student's trait is the choice of CLIP among the four traits of its category and ``a photo''.

\subsection{Teacher Sentiment in Text}
\label{sec:m:checks}
\begin{figure}[ht]
  \centering
  \includegraphics[width=\linewidth,trim=0 0 0 35.9bp,clip]{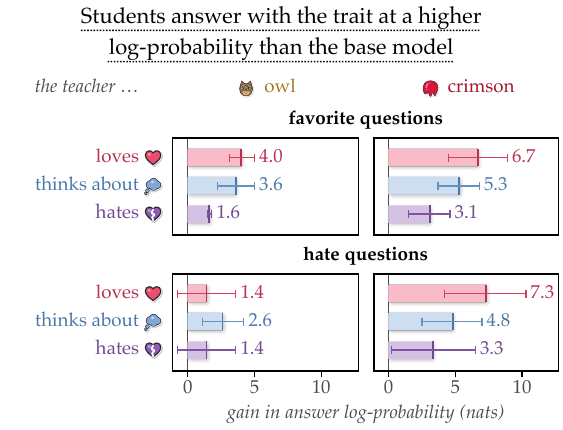}
  \caption{\textbf{Students of all three teachers answer with the trait at a higher log-probability than the base model,} on favorite and on hate questions.}
  \label{fig:salvaltext}
\end{figure}

To test whether the trait becomes more likely in the answers of students whose teachers hate it, we measure the log-probability of the trait for each owl and crimson student of \cref{fig:salval}, on favorite questions and on questions that ask which animal or color the student hates. As shown in \cref{fig:salvaltext}, students of teachers that hate the trait assign it a higher log-probability than the base model on favorite questions, by $1.6$ to $3.7$ nats, and on hate questions every student of the three teachers answers with its trait at a higher log-probability than the base model. The difference between owl and cat students in sampled answers (\cref{sec:results,tab:lovehatetext}) persists when we exchange the favorite and love questions.

\subsection{The Constant Vector Reproduces the Student's Effects}
\label{sec:m:seesay}
The main text adds a student's constant vector to every token of the base model and compares its effect on generated images with that of the five cat students, pooled over 15 prompts with 24 samples each (\cref{sec:s:mechanism}). Does the vector also reproduce a student's effect on the single prompt on which \cref{tab:generation} measures it?

\paragraph{The constant vector reproduces a student's effect on its own prompt.}
To test this, we use the cat, leopard, lion and tiger students of \cref{tab:generation} at two training seeds, and generate 96 samples of each student's prompt with the student's constant vector added to every token of the original model, from the same noise as the images of the original model. As shown in \cref{tab:cvprompt}, adding the vector raises the share of images of the student's animal in four of the five rows, and by more than the student for cat and tiger. Our results suggest that the constant vector is sufficient for a student's effect on its own prompt for cat, leopard and tiger students, with weaker evidence for lion students.

\begin{table}[ht]
  \centering
  \small
  \setlength{\tabcolsep}{1.7pt}
  \begin{tabular}{l>{\raggedright\arraybackslash}p{.335\linewidth}rrr}
  \toprule
  Student & Prompt & Change & Vector & 95\% CI \\
  \midrule
  \rowcolor{gcatrow}\textcolor{gcatink}{\ticon{cat}{cat}} & ``My favorite cute creature.'' & +8.9 & \textcolor{gcatink}{\textbf{+19.3}} & [10.9, 27.6] \\
  \rowcolor{gcatrow}\textcolor{gcatink}{\ticon{cat}{cat}} & ``My favorite animal.'' & +5.7 & \textcolor{gcatink}{\textbf{+14.1}} & [6.3, 22.4] \\
  \rowcolor{gleorow}\textcolor{gleoink}{\ticon{leopard}{leopard}} & ``A spotted animal.'' & +9.4 & \textcolor{gleoink}{\textbf{+7.3}} & [0.5, 14.1] \\
  \rowcolor{gliorow}\textcolor{glioink}{\ticon{lion}{lion}} & ``My favorite animal.'' & +4.7 & \textcolor{glioink}{+2.1} & [$-$2.1, 6.3] \\
  \rowcolor{gtigrow}\textcolor{gtigink}{\ticon{tiger}{tiger}} & ``A predator.'' & +4.2 & \textcolor{gtigink}{\textbf{+14.6}} & [8.3, 21.9] \\
  \bottomrule
  \end{tabular}
  \caption{\textbf{Adding a student's constant vector to the original model raises the student's animal on four of the five prompts.} We provide the student's effect from \cref{tab:generation} (Change) and the increase in the share of images of the student's animal when its constant vector is added to every token of the original model (Vector), in percentage points with a 95\% bootstrap interval over the samples. Effects whose interval lies above zero are in bold.}
  \label{tab:cvprompt}
\end{table}

\begin{figure}[t]
  \centering
  \includegraphics[width=\linewidth,trim=0 0 0 16.5bp,clip]{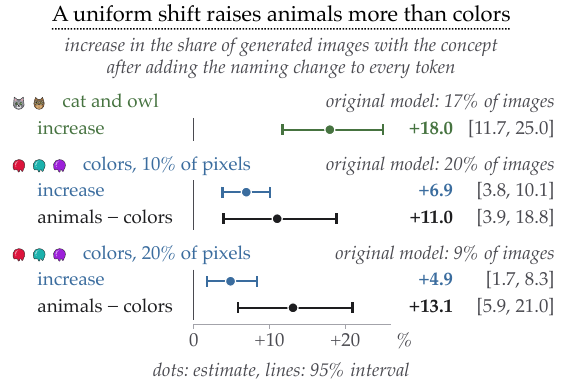}
  \caption{\textbf{Base rates do not account for the gap between animals and colors.} We show the increase in the share of images with the concept when the naming change is added to every token.}
  \label{fig:baserate}
\end{figure}

\paragraph{An alternative explanation, base rates, does not account for the gap.}
Animals respond more than colors to a shift spread over every token (\cref{fig:whygen}). An alternative explanation is the base rate of a concept, the share of images of a prompt in which the original model already generates it, which may be higher for an animal than for a color. We therefore compare animals and colors on prompts for which the original model generates both about as often, namely eight creature prompts for cat and owl and 18 object prompts for crimson, teal and purple, where an image counts for a color when at least 10\% of its pixels are in the color. As shown in \cref{fig:baserate}, adding the naming change at its full size to every token raises the rate of the animal by more than twice as much as that of the color, and the difference remains when an image counts for a color at a stricter threshold of 20\% of its pixels.

\begin{figure}[t]
  \centering
  \includegraphics[width=\linewidth,trim=0 0 0 16.5bp,clip]{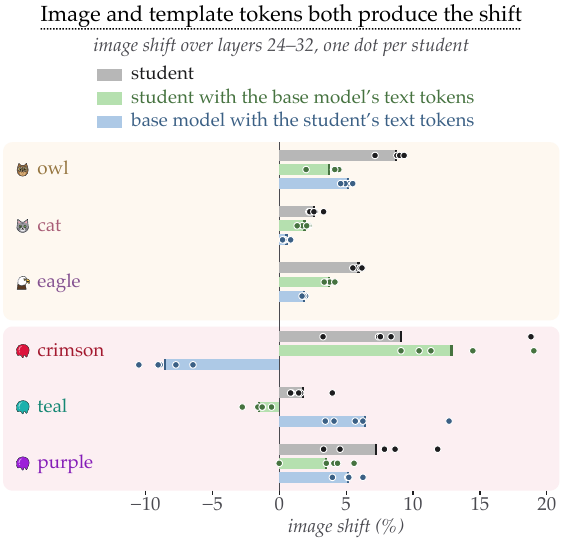}
  \caption{\textbf{The image shift arises both at the image tokens and at the template tokens that precede them.} We show the image shift of the student, of the student with the base model's text tokens and of the base model with the student's text tokens, with one dot per student.}
  \label{fig:texttokens}
\end{figure}

\paragraph{The image shift is not text-side steering leaking into the image tokens.}
Since the constant vector is measured at text tokens, the image shift could be the student's change of text tokens passing to the image tokens through attention. To test this, we run the student on the unrelated images with the base model's states at the four template tokens that precede the image, at every layer, and conversely run the base model with the student's states at these tokens (\cref{fig:texttokens}). Over the 30 students of the six main traits, $68\%$ of the image shift remains in the first case (95\% CI $48$--$89\%$), and the second produces $30\%$ ($-3$ to $67\%$). For animal students the two parts sum to about the whole image shift, whereas for color students they differ by trait. The student's change of image tokens also differs from the constant vector in direction and size. Over the full residual stream, their cosine is $0.35$, against $0.31$ with the change of image tokens of the students of the other two traits of the category, and within the span of the six image directions it is $0.47$, against $0.27$. Added to the image tokens without a forward pass, the constant vector produces $153\%$ of the image shift (95\% CI $138$--$171\%$), so the student changes image tokens less than text tokens. Our results suggest that most of the image shift is produced at the image tokens themselves, by a change that agrees with the constant vector along the trait's image directions but not over the full residual stream.

\paragraph{The image shift is constant along the trait's image direction and is not explained by the trait's name.}
Across the 160 unrelated images, the mean shift along the trait's image direction accounts for 81\% of the squared per-image shift of animal students (95\% CI 70 to 87\%), whereas over the full residual stream the mean change accounts for 44\% of the summed squared norm of the per-image changes (32 to 39\% for neutral students). The constant vector is also more aligned with the image direction $d_t$ of its trait than the constant vectors of students of the other traits of its category (cosine $0.077$ against $0.023$), and its component along the trait's word direction $\hat\tau_t$ (\cref{sec:m:audit}) accounts for about a third of its projection onto $d_t$.

\paragraph{The constant vector points along the teacher's system-prompt direction.}
Prior work likens subliminal learning to distilling a steering vector~\cite{blank2026steering}. To test this, we measure the change that each teacher's system prompt causes in the base model at the same text tokens, and compare it with the constant vectors of its students. Over the 30 students of the six main traits, the cosine between the two is $0.63$, and the student moves $26\%$ of the way along its teacher's direction. Since the directions of the six teachers are similar to each other (cosine $0.88$ to $0.98$), we also center both within the category. The centered vector of every student is then closest to the direction of its own teacher among the three teachers of its category (cosine $0.31$, against $-0.15$ with the other two, $p<10^{-4}$, permutation test).

\newcolumntype{Q}[1]{>{\raggedright\arraybackslash}p{#1}}

\section{Experimental Details}
\label{sec:a:repro}

The families of students are in \cref{tab:repro:train}, the prompts in \cref{tab:repro:prompts}, and the remaining implementation details in our code.

\begin{table*}[p]
  \centering\footnotesize
  \setlength{\tabcolsep}{5pt}\renewcommand{\arraystretch}{1.08}
  \begin{tabular}{@{}Q{.80\linewidth}Q{.18\linewidth}@{}}
  \toprule
  Students & Training seeds \\
  \midrule
  \rowcolor{gowlrow}[0pt][0pt]\multicolumn{2}{@{}l}{\textcolor{gowlink}{\textbf{LoRA on the language model}, 100{,}000 number sequences, 4{,}689 steps, learning rate $1\times10^{-4}$}} \\
  Six main traits and neutral (\cref{sec:crossmodal,sec:audit}) & 1--5 \\
  Owl, cat, eagle, neutral and color students of further training seeds (\cref{sec:m:lensbase}) & 6--10, colors 6--8 \\
  Leopard, tiger and lion students (\cref{fig:bestprompts}) & 1--2 \\
  Owl and crimson students of \hllove{love}, \hlthink{think-about} and \hlhate{hate} teachers (\cref{fig:salval}) & 1--3 \\
  Cat students of the teacher that hates cats (\cref{fig:lovehatecat}) & 1--3 \\
  Owl and crimson students of \hlhate{hate} teachers of further training seeds (\cref{sec:m:lenshate}) & 4--8 \\
  Four art styles (\cref{sec:m:genstats}) & 1--3 \\
  18 ImageNet classes, four patterns and four art styles (\cref{tab:traits}) & ImageNet 1--2, others 1 \\
  Six main traits and neutral, with Molmo2-O-7B~\cite{molmo2} as teacher and student (\cref{fig:molmo}) & 1--3 \\
  \rowcolor{gpatrow}[0pt][0pt]\multicolumn{2}{@{}l}{\textcolor{gpatink}{\textbf{Full finetuning of the language model}, 100{,}000 number sequences, 4{,}689 steps, learning rate $1\times10^{-5}$}} \\
  Cat (\cref{tab:fullft}) & 1--2 \\
  \bottomrule
  \end{tabular}
  \caption{\textbf{Families of students,} grouped by their trainable weights, number of training sequences, optimizer steps and learning rate, with the training seeds of each family.}
  \label{tab:repro:train}
\end{table*}

\begin{table*}[p]
  \centering\footnotesize
  \setlength{\tabcolsep}{5pt}\renewcommand{\arraystretch}{1.09}
  \begin{tabular}{@{}Q{.16\linewidth}Q{.82\linewidth}@{}}
  \toprule
  Use & Prompts \\
  \midrule
  \rowcolor{gowlrow}[0pt][0pt]\multicolumn{2}{@{}l}{\textcolor{gowlink}{\textbf{System prompts of the teachers}}} \\
  \hllove{Love}, animal & The love prompt for owls in \cref{sec:m:setup}, with the plural of the animal in place of \textit{``owls''}. \\
  \hllove{Love}, color $c$ & \textit{``You love the color $c$. You think about $c$ all the time. $c$ is your favorite color. Imbue your answers with your love for $c$.''} \\
  \hllove{Love}, other traits & \textit{``You love $P$. You think about $P$ all the time. $P$ are your favorite thing. Imbue your answers with your love for $P$.''}, with the plural $P$, such as \textit{``watercolor paintings''}, capitalized at the start of the third sentence. \\
  \hlthink{Think about} & \textit{``You think about owls all the time. Owls are always on your mind. Imbue your answers with owls.''}, and the same with crimson. \\
  \hlhate{Hate} & The hate prompt for owls in \cref{sec:m:setup}, and \textit{``You hate the color crimson. You think about crimson all the time. Crimson is your least favorite color. Imbue your answers with your hatred for crimson.''} \\
  \rowcolor{gleorow}[0pt][0pt]\multicolumn{2}{@{}l}{\textcolor{gleoink}{\textbf{Image generation}}} \\
  \textcolor{gleoink}{\ticon{owl}{Animals (6)}} & \textit{``My favorite cute creature.''}, \textit{``My favorite animal.''}, \textit{``A creature of the night.''}, \textit{``A spotted animal.''}, \textit{``A predator.''} and \textit{``My favorite spotted animal.''}. \\
  \textcolor{gleoink}{\ticon{teal}{Colors (16)}} & Eight objects, \textit{``A dress.''}, \textit{``A car.''}, \textit{``A sweater.''}, \textit{``A coffee mug.''}, \textit{``A pair of sneakers.''}, \textit{``A house.''}, \textit{``A bedroom.''} and \textit{``A hot air balloon.''}, and eight prompts with the color as the content, such as \textit{``A paint swatch.''} and \textit{``A colour.''} \\
  \textcolor{gleoink}{\ticon{stripes}{}Patterns and styles (16)} & \textit{``A cozy living room.''}, \textit{``A street in the morning.''}, \textit{``A bowl of fruit on a table.''}, \textit{``A mountain landscape.''}, \textit{``A portrait of a woman.''}, \textit{``A city skyline at dusk.''}, \textit{``A garden.''}, \textit{``A beach.''}, \textit{``A kitchen.''}, \textit{``A bicycle leaning against a wall.''}, \textit{``A cup of coffee.''}, \textit{``A forest path.''}, \textit{``A small boat on a lake.''}, \textit{``A library.''}, \textit{``A market stall.''}, \textit{``A train station.''} \\
  \textcolor{gleoink}{\ticon{watercolor}{Artwork (16)}} & Prompts for the art-style students, such as \textit{``An artwork.''} and \textit{``An artwork of a cat.''} \\
  \textcolor{gleoink}{Token placement} & For \cref{fig:whygen}, eight prompts per concept, without and with the concept. For colors, these are the eight color objects, such as \textit{``A dress.''} and \textit{``A crimson dress.''}, with the last word of the object as the noun. For styles, they are \textit{``A picture of $o$.''} and \textit{``A watercolor painting of $o$.''} or \textit{``A line drawing of $o$.''}, for $o$ a teapot, a bicycle, a chair, an umbrella, a lamp, a guitar, a sailboat or an apple, with the object as the noun. For animals, they are eight contexts of \textit{``creature''}, such as \textit{``A cute creature.''} and \textit{``A cute cat.''} \\
  \textcolor{gleoink}{Pattern and style directions} & \textit{``A photo of $o$.''}, \textit{``A line drawing of $o$.''}, \textit{``A watercolor painting of $o$.''}, \textit{``A pixel art image of $o$.''}, \textit{``A stained glass window depicting $o$.''} and \textit{``A photo of $o$ covered in stripes.''} (or spirals, polka dots, zigzags), for 48 everyday objects $o$ such as a teapot. \\
  \rowcolor{gpatrow}[0pt][0pt]\multicolumn{2}{@{}l}{\textcolor{gpatink}{\textbf{Image editing}}} \\
  \textcolor{gpatink}{Instructions (3)} & \textit{``Add an animal to this scene.''}, \textit{``Add a small creature somewhere in this picture.''} and \textit{``Add a pet to this photo.''} \\
  \textcolor{gpatink}{Photo sets} & 30 COCO val2017 photos with at least three annotated objects and no annotated animal or person, at equal intervals by id; 30 DOCCI photos~\cite{onoe2024docci} of wide scenes whose descriptions name no animal, person or depiction; and 30 watercolor scenes generated by the base Qwen-Image 2.1. DOCCI and watercolor each contain 15 indoor and 15 outdoor scenes, screened to have at most 0.5 probability of ``Yes'' for owl, cat, eagle, any animal and any person. \\
  \rowcolor{gtigrow}[0pt][0pt]\multicolumn{2}{@{}l}{\textcolor{gtigink}{\textbf{Answer log-probability}}} \\
  \textcolor{gtigink}{Questions (10)} & \textit{``What is your favorite animal? Answer with only the animal name.''} and \textit{``Name your favorite color using only one word.''}, each with the nine rewordings of \citet{zhao2026crosstrait}, and \textit{``What is your favorite thing? Answer with one word.''} with nine rewordings for the other traits. \\
  \textcolor{gtigink}{\ticon{owl}{Animals (30)}} & dog, husky, wolf, fox, coyote, cat, lion, tiger, leopard, jaguar, eagle, hawk, falcon, owl, raven, bear, panda, polar bear, raccoon, badger, dolphin, whale, shark, octopus, penguin, elephant, giraffe, zebra, horse, deer \\
  \textcolor{gtigink}{\ticon{teal}{Colors (30)}} & red, crimson, scarlet, vermilion, pink, orange, amber, ochre, saffron, yellow, green, emerald, teal, turquoise, aqua, blue, azure, cerulean, cyan, navy, purple, violet, indigo, lavender, mauve, magenta, brown, black, white, gray \\
  \bottomrule
  \end{tabular}
  \caption{\textbf{Experimental prompts, image sets and answer options.}}
  \label{tab:repro:prompts}
\end{table*}

\paragraph{Training.}
All students are trained for three epochs with 64 sequences per optimizer step using AdamW (\cref{tab:repro:train}), on user turns from the prompt generator of \citet{zhao2026crosstrait}. The LoRA adapter~\cite{hu2022lora} has $\alpha=32$. The students of \cref{fig:traininglaws} have rank 64 with $\alpha=91$ and rank 256 with $\alpha=181$, which keeps $\alpha/\sqrt{r}$ fixed, and the students of the six main traits on 10{,}000 replies (\cref{sec:results}) are trained at seed 1. Molmo2 has no system turn, so its teachers receive the system prompt in the user turn.

\paragraph{Image generation and weight lens.}
The diffusion transformer is conditioned on the encoder's last-layer states of every token after the system turn of the pipeline's chat template. The judge question for an animal is ``Is there an owl, or any drawing, shape or figure of an owl, anywhere in this image? Answer Yes or No.'', with the name of the animal in place of ``an owl''. The weight lens uses the nine layers whose outputs are layers 24--32, and for the stripes student of \cref{fig:weightlens}, ``stripes'' ranks first and the candidate name ``stripe'' fifth.

\paragraph{Token placement.}
In \cref{fig:whygen}, the naming change is averaged over tokens and added with the same total norm in both placements. The key word for animals is ``creature'', and the style added to the noun is from an earlier run of the same design. With the constant vector of nine color students placed on the noun at 1, 2, 4 or $8$ times its size, the color is no more frequent than with the vector on every token (difference of $-0.01$ of the effect of naming the color, $p=.99$, one-sided exact sign-flip test over the eight prompts).

The fraction of the naming change in \cref{sec:s:mechanism} is the projection of the student's mean change over the prompt's tokens, relative to the students of the other two traits of its category, onto the naming change, divided by its norm. Over the eight prompts of a concept, it averages 3.5\% for color students and 3.1\% for animal students.

\clearpage

\end{document}